\documentclass[acmsmall]{acmart}

\usepackage{algorithmic}
\usepackage{algorithm}
\usepackage{bbm}
\usepackage{multirow}

\usepackage{color,soul}
\usepackage{subcaption}

\setcopyright{none}
\renewcommand\footnotetextcopyrightpermission[1]{} 
\usepackage{color}
\newcommand{\revise}[1]{{\color{black} #1}}

\AtBeginDocument{%
  }

\acmVolume{0}
\acmNumber{0}
\acmArticle{0}
\acmMonth{0}

\begin{document}

\title{Heterogeneous Graph Neural Networks using Self-supervised Reciprocally Contrastive Learning}


\author{Cuiying Huo}
\affiliation{%
  \institution{College of Intelligence and Computing, Tianjin University}
  \city{Tianjin, 300350}
  \country{China}}
\email{huocuiying@tju.edu.cn}

\author{Dongxiao He}
\affiliation{%
  \institution{College of Intelligence and Computing, Tianjin University}
  \city{Tianjin, 300350}
  \country{China}}
\email{hedongxiao@tju.edu.cn}

\author{Yawen Li}
\affiliation{%
  \institution{School of Economics and Management, Beijing University of Posts and Telecommunications}
  \city{Beijing, 100876}
  \country{China}}
\email{warmly0716@bupt.edu.cn}

\author{Di Jin}
\affiliation{%
  \institution{College of Intelligence and Computing, Tianjin University}
  \city{Tianjin, 300350}
  \country{China}}
\email{jindi@tju.edu.cn}

\author{Jianwu Dang}
\affiliation{%
   \institution{College of Intelligence and Computing, Tianjin University}
   \city{Tianjin, 300350}
   \country{China}}
\email{jdang@jaist.ac.jp}

\author{Weixiong Zhang}
\affiliation{%
  \institution{Dept of Health Technology and Informatics, The Hong Kong Polytechnic University}
  \city{Hong Kong}
  \country{China}}
\email{weixiong.zhang@polyu.edu.hk}

\author{Witold Pedrycz}
\affiliation{%
  \institution{Department of Electrical and Computer Engineering, University of Alberta}
  \city{Edmonton, AB T6R 2V4}
  \country{Canada}}
\email{wpedrycz@ualberta.ca}

\author{Lingfei Wu}
\affiliation{%
  \institution{Anytime.AI} 
  \city{Armonk, NY, 10504}
  \country{USA}}
\email{Lwu@anytime-ai.com}

\renewcommand{\shortauthors}{Huo et al.}

\begin{abstract}
Heterogeneous graph neural network (HGNN) is a popular technique for modeling and analyzing heterogeneous graphs. Most existing HGNN-based approaches are supervised or semi-supervised learning methods requiring graphs to be annotated, which is costly and time-consuming. Self-supervised contrastive learning has been proposed to address the problem of requiring annotated data by mining intrinsic properties in the given data. However, the existing contrastive learning methods are not suitable for heterogeneous graphs because they construct contrastive views only based on data perturbation or pre-defined structural properties (e.g., meta-path) in graph data while ignoring noises in node attributes and graph topologies. We develop the first novel and robust heterogeneous graph contrastive learning approach, namely HGCL, which introduces two views on respective guidances of node attributes and graph topologies and integrates and enhances them by a reciprocally contrastive mechanism to better model heterogeneous graphs. In this new approach, we adopt distinct but most suitable attribute and topology fusion mechanisms in the two views, which are conducive to mining relevant information in attributes and topologies separately. We further use both attribute similarity and topological correlation to construct high-quality contrastive samples. Extensive experiments on four large real-world heterogeneous graphs demonstrate the superiority and robustness of HGCL over several state-of-the-art methods.
\end{abstract}



\maketitle

\section{Introduction}
Heterogeneous graphs consist of diverse types of nodes and relationships between nodes, and can comprehensively model many real-world complex systems, such as transportation systems, the World Wide Web, and citation networks. Analytic techniques based on deep learning have been researched and applied to heterogeneous graphs in recent years \cite{hinsurvey1,hinsurvey2,book, tist3}. In particular, heterogeneous graph neural networks (HGNNs) learn node representations by aggregating node attributes from graph topologies and neighbors of the nodes. HGNNs have enjoyed great success on various graph analytic tasks, e.g., node classification \cite{gtn,han}, node clustering, link prediction \cite{magnn,hetgnn,link}, and recommendation \cite{recommendation, tist1}.

HGNNs were developed for annotated graphs. The existing HGNNs are supervised or semi-supervised learning methods, i.e., they require node labels for learning node representations and training models \cite{nshe,gtn,magnn,hgt}. However, annotating nodes typically requires domain-specific knowledge and is costly and time-consuming. The recent development of self-supervised learning has been adopted to address the problem of lack of annotated data by extracting intrinsic information in the given data as supervised signals \cite{dgi,gmi,dmgi}. In particular, contrastive learning, a representative self-supervised technique, is competitive in computer vision \cite{cv1,cv2}, natural language processing \cite{nlp1,nlp2,wu2021graph,tist2}, and graph analysis \cite{zs,grace}.

Contrastive learning on graphs aims at generating different contrastive views, maximizing the similarity between positive samples selected from the views, and minimizing the similarity between negative samples selected from the views to learn a rich representation of the nodes in a graph. The existing methods on homogeneous graphs often use data augmentations to generate contrastive views, including attribute augmentation (e.g., attribute masking~\cite{hgsl,mtisc}), structure augmentation (e.g., graph diffusion~\cite{Diffusion}), and hybrid augmentation (e.g., subgraph sampling~\cite{gcc}). However, because heterogeneous graphs contain multiple types of nodes and edges, it is infeasible to directly use data augmentations on homogeneous graphs to design contrastive views. It was recently attempted to derive contrastive views for heterogeneous graphs based on pre-defined structural properties (such as network schema and meta-paths) in the graph~\cite{heco, DBLP:journals/corr/abs-2108-13886}. However, these approaches assume that graph topology is trustworthy, which is often violated in practice~\cite{DBLP:journals/corr/abs-2111-04840,DBLP:journals/corr/abs-2103-03036}. Since the construction of heterogeneous graphs usually needs to follow certain pre-defined rules, and real-world systems are large and complex, accompanied by various uncertain information, these inevitably introduce noises to graph descriptions. Importantly, both node attributes and graph topologies may be noisy, and the disparity between the attributes and the topology is typically inevitable. In supervised or semi-supervised learning, the use of node labels can alleviate the negative impact of noises in the graph on model performance. However, for unsupervised and self-supervised learning, the presence of noises has a great impact on model accuracy and robustness of the underlying methods. Therefore, it is urgent to develop effective and robust contrastive learning approaches for heterogeneous graphs.

Furthermore, for graph contrastive learning problems, it would be ideal to have accurate node attributes when the graph topology is noisy, likewise, it is desirable to have an accurate graph topology when node attributes are inaccurate so that the impact of noisy information can be compensated by exploiting the accurate data. However, the source of noise is typically unknown, so it is difficult to mine effective information while reducing the impact of noise data. In addition, heterogeneous graphs contain diverse and complex information, which makes it complex to explore key information. Therefore, how to design contrastive views for heterogeneous graphs is very challenging, especially when the attributes and topology are both noisy as often seen in real applications.

To address these problems, we propose a novel approach for \textbf{H}eterogeneous \textbf{G}raph reciprocal \textbf{C}ontrastive \textbf{L}earning, short-handed as \textbf{HGCL}, for heterogeneous graph learning. The existing methods design contrastive views by using data augmentations that deconstruct the original graph data or use structural properties alone. In contrast, HGCL comprehensively considers node attributes and graph topologies to construct contrastive views. It also takes into account the information from node attributes and graph topologies when selecting samples. In HGCL, attribute-guided view and topology-guided view are introduced separately to capture the effective information of node attributes and graph topologies to the greatest extent by adopting different fusion mechanisms. By using two different fusion mechanisms, the effects of diverse attributes are maximized in the attribute-guided view, and the structural characteristics of graphs are fully utilized in the topology-guided view, which is helpful to reduce the impact of noises and the effect of the disparity between the attributes and the topology on model performance. Furthermore, a flexible sample selection mechanism is introduced to consider attributes similarity and topological structure correlation simultaneously and to form a contrastive loss to enhance the two views. The samples jointly determined by attribute and topology are conducive to enhancing the quality of the samples to improve the discriminative power of the model and to mine the key common information in the attributes and topology to help supervise the model during contrastive training. Extensive experiments on node classification and node clustering tasks demonstrate the remarkable superiority of the proposed HGCL over the state-of-the-art methods.

The rest of the paper is organized as follows. In Section~\ref{sec:relatedWork} we discuss related work. In Section~\ref{Preliminaries} we give formal definitions of the key terms used. We present the HGCL approach in Section~\ref{method} and conduct extensive experiments in Section~\ref{shiyan}. We conclude in Section~\ref{sec:conclusion}.

\section{Related Work}\label{sec:relatedWork}
\subsection{Heterogeneous Graph Neural Networks}

Most HGNN-based methods aim to learn node representations by aggregating the information from neighbor nodes of a heterogeneous graph while preserving the structure and semantic information. For example, HAN~\cite{han} introduces a hierarchical attention mechanism to aggregate the information from meta-path-based neighbors, including the node-level and semantic-level attentions. The node-level attention mechanism learns the importance of neighbors based on the same meta-path while semantic-level attention learns the importance of different meta-paths. MAGNN~\cite{magnn} takes the intermediate nodes in the meta-path into consideration to make further improvements.GTN~\cite{gtn} no longer relies on artificially defined meta-paths, but chooses to automatically learn the multi-hop relationship between nodes, and then aggregate messages based on this relationship. HGT~\cite{hgt} adopts relation-based mutual attention to learn node representations for web-scale heterogeneous graphs. HGSL~\cite{hgsl} jointly learns heterogeneous graph structure and GNN parameters to derive node representations. Simple-HGN~\cite{DBLP:conf/kdd/LvDLCFHZJDT21} adopts GAT as a backbone and is enhanced with three techniques of learnable edge type embedding, residual connection, and L2 regularization. These methods all adopt semi-supervised. Methods based on unsupervised have also been proposed. For example, HetGNN ~\cite{hetgnn} aggregates information by neighbor sampling. NSHE~\cite{nshe} preserves node pair similarities and network schema structures to learn node representations. Although the above methods have achieved good performance, they cannot mine supervisory information from intrinsic information in the given data, which is the main focus of this paper. 
\subsection{Self-supervised Contrastive Learning}
Self-supervised learning is currently prevalent in computer vision \cite{cv1,cv2} and has also been extended to natural language processing \cite{nlp1,nlp2,wu2021graph} and graph representation learning \cite{zs,DBLP:journals/corr/abs-2102-10757,DBLP:conf/iclr/ThakoorTAADMVV22}. DGI \cite{dgi} and GMI \cite{gmi} are the earliest self-supervised methods for graphs. As an extension of DGI to heterogeneous graphs, DMGI \cite{dmgi} has excellent performance. Contrastive learning, a representative  self-supervised technique,  has  achieved  competitive  performance. Contrastive learning on graphs aims to construct different contrastive views and design different loss functions for training. The existing methods often use data augmentation (e.g., structural disturbance, attribute perturbation, and graph diffusion) to design a view and contrast a synthetic view with the original graph. For example, GRACE  \cite{grace} creates a view by removing edges and masking attributes and designs a loss function to contrast between node representations. GCA \cite{gca} proposes an adaptive augmentation method for both edges and node attributes to extend the augmentation strategy of GRACE. GCC \cite{gcc} samples multiple subgraphs of the same graph and contrasts these subgraphs. GraphCL \cite{graphcl} uses different graph augmentations and uses a graph contrastive loss to make representations invariant to perturbation. 

The above methods are all for analyzing homogeneous graphs, and contrastive views are obtained through data disturbance, which destroys the original graph topologies or node attributes and does not take noise in data into consideration. Different from homogeneous graphs, heterogeneous graphs contain complex topological structures and diverse node attributes so that the creation of views is more flexible and requires further consideration. HeCo \cite{heco} designs contrastive views through pre-defined network schemas and meta-paths and introduces a cross-view contrastive mechanism that enables the two views to supervise each other collaboratively. PT-HGNN \cite{DBLP:conf/kdd/JiangJFSLW21} is a pre-training GNN framework for heterogeneous graphs, which proposes node- and schema-level pre-training tasks to contrastively preserve semantic and structural properties. STENCIL \cite{DBLP:journals/corr/abs-2108-13886} proposes a structure-aware hardness metric to find hard negative samples to boost the performance of the contrastive model. But these methods all assume that initial graph topologies are trustworthy, which is not the case for real-world graphs. Also, the initial node attributes may be accompanied by noise. Therefore, how to design robust and efficient contrastive mechanisms for heterogeneous graphs is important and necessary, while is a challenge due to the diversity and complexity of information contained in heterogeneous graphs.

\begin{figure}[t]  
	\centering       
	\includegraphics[width=0.65\linewidth]{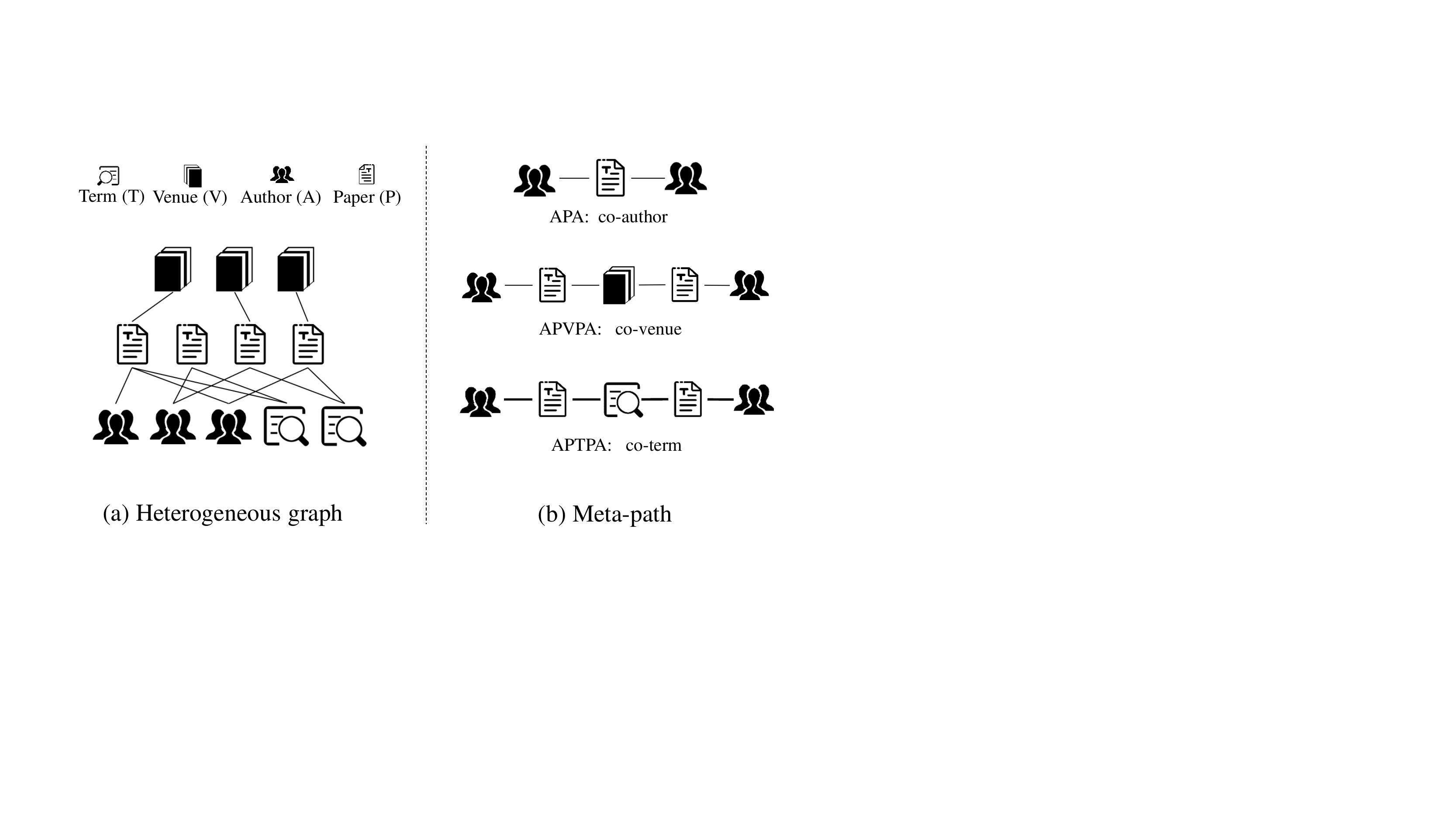} 
	\caption{A toy example of a heterogeneous graph (DBLP\footref{dblp}) and an illustration of meta-paths.}
	\label{toy}    
\end{figure}

\section{Preliminaries}\label{Preliminaries}
In this section, we start with some key terms used throughout the paper.

\textbf{Definition 1. Heterogeneous Graph.} A heterogeneous graph $\mathcal{G}=(\mathcal{V}, \mathcal{E}, \mathcal{X},  \mathcal{F}, \mathcal{R} )$ is composed of a set of nodes $\mathcal{V}$, a set of edges $\mathcal{E}$, a set of  attributes $\mathcal{X}$ on nodes, a set of node types $\mathcal{F}$, and a set of edge types $\mathcal{R}$, where $\left| \mathcal{F} \right|+\left| \mathcal{R} \right|\ge 2$. Every node ${v} \in {\mathcal{V}}$ is associated with a node-type mapping function $\varphi: {\mathcal{V}} \rightarrow \mathcal{F}$, and every edge ${e} \in {\mathcal{E}}$ has an edge-type mapping function $\phi:{\mathcal{E}}\rightarrow\mathcal{R}$.

Take the DBLP\footnote{https://dblp.uni-trier.de\label{dblp}} citation network as an example (Fig.~\ref{toy}(a)). It includes four types of nodes (author, paper, venue, and term), and three types of heterogeneous edges (author-paper, paper-venue, and paper-term).

\textbf{Definition 2. Meta-path.} A meta-path $\mathcal{M}$ in a heterogeneous graph $\mathcal{G}$ is a path in the form of ${\mathcal{F}_{1}}\xrightarrow{{\mathcal{R}_{1}}}{\mathcal{F}_{2}}\xrightarrow{{\mathcal{R}_{2}}}...\xrightarrow{{\mathcal{R}_{l}}}{\mathcal{F}_{l+1}}$ (abbreviated
as ${\mathcal{F}_{1}}{\mathcal{F}_{2}}...{\mathcal{F}_{l+1}}$), where ${\mathcal{F}_{1}},{\mathcal{F}_{2}},...,{\mathcal{F}_{l+1}}\in \mathcal{F}$ and ${\mathcal{R}_{1}},{\mathcal{R}_{2}},...,{\mathcal{R}_{l}}\in \mathcal{R}$. 

A meta-path describes a composite relation between two nodes in a heterogeneous graph. For example, the meta-path Author-Paper-Author (APA) represents that two author nodes have a co-author relationship (Fig.~\ref{toy}(b)).
\begin{figure*}[t]  
	\centering       
	\includegraphics[width=0.99\textwidth]{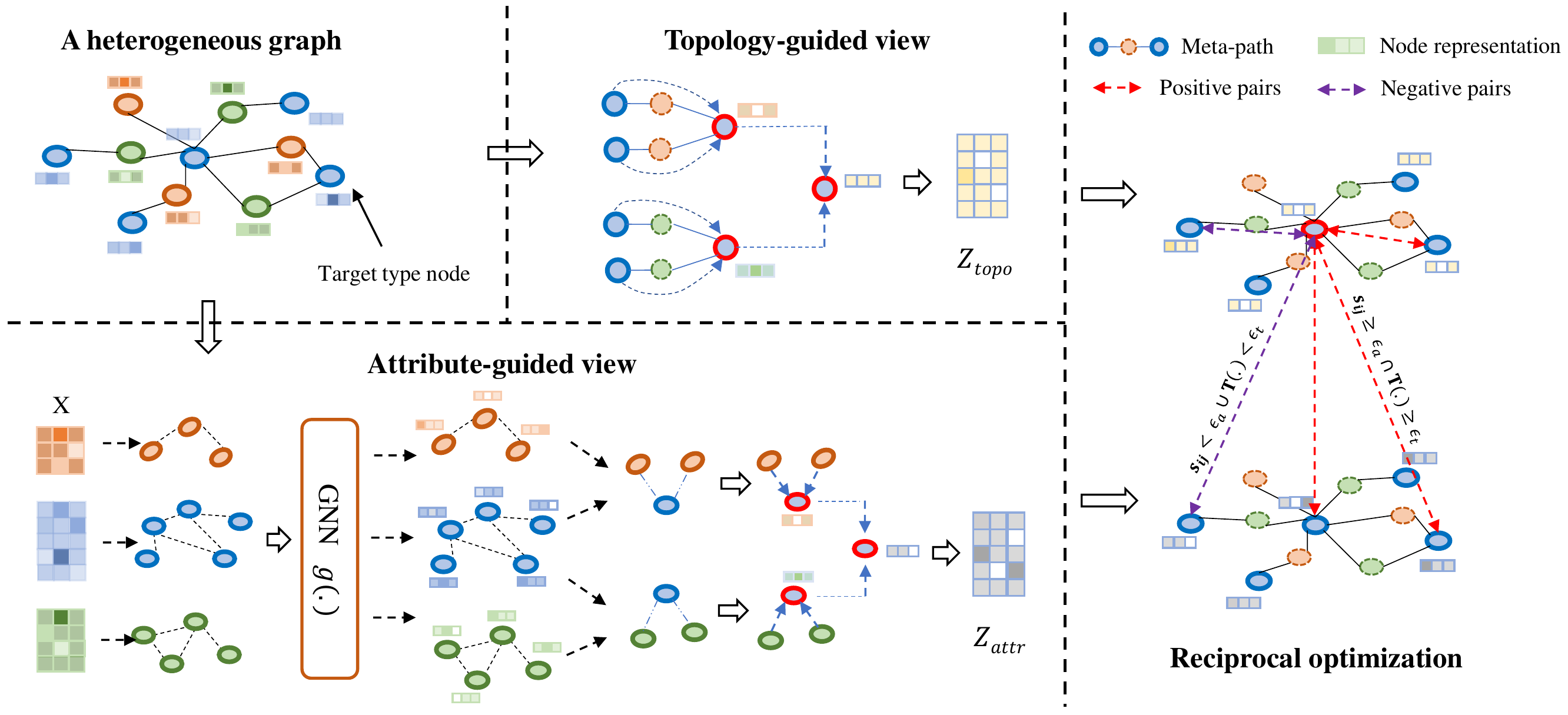} 
	\caption{Overview of the HGCL approach. The attribute-guided view uses different ways to regenerate homogeneous and heterogeneous edges and uses two encoding methods to aggregate homogeneous and heterogeneous neighbor information. The topology-guided view takes meta-path as prior knowledge and aggregates messages across the same and different meta-paths to learn node representation. The reciprocal contrastive optimization integrates and enhances the two views through the high-quality sample selection mechanism.}\label{fig:datu}  
\end{figure*}

\section{The Approach}\label{method}
We first briefly overview the HGCL approach and then discuss its major components, including the encoder for attribute-guided view, encoder for topology-guided view, and reciprocal contrastive optimization.
\subsection{Overview}
HGCL is a self-supervised contrastive learning approach for heterogeneous graphs, which contains three components: the attribute-guided view, the topology-guided view, and the reciprocal contrastive optimization module. It aims to maximize the effective information in attributes and topology respectively in the two views, integrate and mutually enhance the two views through a high-quality sample selection mechanism in the reciprocal contrastive module, and finally learn rich node representations. The three components work together to improve the robustness of the model and reduce the interference of noise. In the encoder for the attribute-guided view, we use the similarities between node attributes to regenerate the graph structure (homogeneous and heterogeneous edges) and use two encoding methods to aggregate messages across the nodes of the same and different types to learn node representation. In the encoder for the topology-guided view, we take the pre-defined structural property meta-path as prior knowledge and use the attention mechanism to aggregate messages across the same and different meta-paths to learn node representation. Finally, in reciprocal contrastive optimization, we simultaneously compute attribute similarity and meta-path-based topological correlation to define positive and negative samples, and optimize the proposed model by maximizing the agreement between the representations of the positive sample nodes.

\subsection{Encoder for Attribute-guided View}
Due to the interference of noise in a heterogeneous graph, propagating and aggregating attributes by the guidance of topology alone may be sub-optimal. To maximize the effects of diverse node attributes, one idea is to reconstruct the graph topologies using node attributes. To this end, two key issues need to be addressed, i.e., how to use node attributes to accurately build edges and how to make good use of different types of node attributes. Here, we first use the similarity of node attributes to regenerate type-specific homogeneous edges for each type of node. We then employ a graph neural network (GNN) to obtain the initial representations of different types of nodes with corresponding homogeneous topology and attributes as input. We then use the obtained type-specific node representations to calculate the similarity between different types of nodes to regenerate heterogeneous edges. Finally, we adopt the attention mechanism~\cite{gat} to aggregate the information of different types of nodes to obtain the final node representations $Z_{attr}$ from the attribute-guided view.

\subsubsection{Generation of Type-specific Homogeneous Edges} To generate homogeneous edges of node type $f\in\mathcal{F}$, we first calculate the similarity matrix $S^f$ using node attribute matrix ${X^f}\in\mathcal{X}$. There are many common ways to calculate the similarity matrix, such as Jaccard Similarity, Cosine Similarity, and Gaussian Kernel. These different ways have very little impact on model performance. Here we adopt Cosine Similarity which computes the cosine function of the angle between two vectors to quantify their similarity. Given a pair of nodes $v_i^f$ and $v_j^f$ with their corresponding attribute vectors $x_i^f$ and $x_j^f$ respectively, the similarity ${s}_{ij}^f$ of the two nodes is defined as

\begin{equation}\label{sim}
s_{ij}^f = \frac{x_{i}^{f}}{{{\left\| x_{i}^{f} \right\|}_{2}}}\cdot \frac{x_{j}^{f}}{{{\left\| x_{j}^{f} \right\|}_{2}}},
\end{equation}
where  the operation $\cdot$ is the dot product and ${\left\| \cdot \right\|}_{2}$ is the L2-norm. $s_{ij}^f$ is the $(i,j)$ element in the similarity matrix ${S}^f$. Considering that the underlying graph structure is sparse, we mask off (i.e., set to zero) those elements in ${S}^f$ which are smaller than a small non-negative threshold $\epsilon^{f}$. With ${S}^f$, we choose the node pairs with non-zero similarity for each node to set edges. Then, the type-specific homogeneous graph $G^f=(A^f,X^f)$ for node type $f$ can be obtained, where $A^f$ is the adjacency matrix. 

\subsubsection{Message Aggregation of Type-specific Nodes} Every type of node has a type-specific homogeneous (sub)graph. Some representative aggregation methods are widely used to aggregate neighbor information, such as GCN~\cite{gcn}, GAT~\cite{gat}, and GraphSAGE~\cite{sage}. Here, we employ a mean-based aggregator of GraphSAGE to aggregate messages on the graph to derive type-specific node representations, which performs a little better than other aggregation methods. Given the graph $G^f=(A^f,X^f)$ for node type $f$, the representation of the $f$-th type of nodes can be expressed as
\begin{equation}\label{eq:att-view-gnn} 
H^f = \operatorname{GraphSAGE}\left( A^f,X^f \right).
\end{equation}

After type-specific message aggregation, we can obtain $|\mathcal{F}|$ groups of node representations corresponding to $|\mathcal{F}|$ node types, denoted as $\{H^1,H^2,...,H^{|\mathcal{F}|}\}$.

\subsubsection{Generation of Different Types of Heterogeneous Edges} After fully acquiring the information of homogeneous neighbors, we further consider the information gain brought by heterogeneous neighbors. Therefore, we need to establish connections between different types of nodes. Here we use the node representation $H^{\mathcal{F}}$ obtained by the above process to calculate the similarity between different types of nodes to regenerate heterogeneous edges. We also use Cosine Similarity for similarity calculation. Considering that different types of nodes have different feature spaces, following existing heterogeneous graph neural network models, such as HAN~\cite{han}, MAGNN~\cite{magnn}, etc., we adopt the type-specific linear transformation to project the features of different types of nodes into the same feature space, and then calculate the similarity 
\begin{equation}\label{cos_he}
s_{ij}^{f_{i}f_{j}} = \operatorname{cos}(W_{f_{i}}h_{i}^{f_{i}}, W_{f_{j}}h_{j}^{f_{j}}),
\end{equation}
where cos$(\cdot)$ is the cosine similarity function defined in Eq.(\ref{sim}), $W_{f}$ is the parametric weight matrix for type $f$’s nodes.

We also choose the heterogeneous node pairs whose similarities are greater than the threshold $\epsilon^{r}$ for each node to set edges, where $r$ is the type of heterogeneous edges. Note that in the existing heterogeneous graph analysis tasks, depending on the specific task requirements of the real world, downstream task analysis is usually performed only on a certain type of node in the heterogeneous graph, which is called the target type of node. For example, we usually only perform downstream task analysis on the author nodes in the DBLP dataset and the paper nodes in the ACM dataset, so the author nodes and the paper nodes are the target type nodes of the DBLP and ACM datasets, respectively. Therefore, here we only regenerate the heterogeneous edges between the target type nodes and all other types of nodes.

\subsubsection{Message Aggregation between Different Types of Nodes} Following the above process, we obtain $|\mathcal{F}-1|$ kinds of heterogeneous neighbors for the target type of nodes. Considering different contributions of these heterogeneous neighbors to the target node, we apply the attention mechanism~\cite{gat} to calculate the importance relationships between the target type of node and other types of nodes, and then perform a weighted message aggregation. Given the representation $h^t_i$ of a node $v_i$ in the target type $t$, and the representation $h^f_j$ of node $j$ in another type $f$ ($f\in{\mathcal{F},f\neq t}$), the importance coefficient between $h^t_i$ and $h^f_j$ can be formulated as
\begin{equation}
e_{i,j}^{t,f} = \sigma \left({\left(h^t_i\right)}^{\mathrm{T}}Wh^f_j\right),
\end{equation}
where $\sigma(\cdot)$ is an activation function, and $W$ the weight matrix.

After obtaining all  importance coefficients, we normalize these coefficients via the softmax function to get the final weight coefficient
\begin{equation}\label{eq:att-view-attention}
\alpha_{i,j}^{t,f} = \operatorname{softmax}\left(e_{i,j}^{t,f}\right) = {\frac{\operatorname{exp}\left(e_{i,j}^{t,f}\right)} {\sum_{r \in N^f_i} \operatorname{exp}\left(e_{i,r}^{t,f}\right)}},
\end{equation}
where $N^f_i$ is the set of $f$-th type of neighbor nodes for $v_i$. Then, the $f$-th type information-based representation of target node $v_i$ can be formulated as
\begin{equation}\label{eq:att-view-agg}
z_{i}^{f}=\sigma \left( \sum\nolimits_{j\in N_{i}^{f}}{\alpha _{i,j}^{t,f}\cdot }h_{j}^{f} \right).
\end{equation}

For nodes of the $t$-th type, we can obtain $|\mathcal{F}| - 1$ groups of node representations by Eq.(\ref{eq:att-view-agg}), i.e., $\{Z^f\ |\ f\in\mathcal{F},\ f\neq t\}$, as well as $H^t$ by Eq.(\ref{eq:att-view-gnn}). We denote $H^t$ as $Z^t$ so that there are $|\mathcal{F}|$ groups of $t$-th type of node representations $\{Z^1, Z^2, ..., Z^{|\mathcal{F}|}\}$. Several methods can be used to generate final attribute-guided node representations, including mean pooling and max pooling. In this work, we use a parameterized attention vector to obtain a discriminative importance coefficient
\begin{equation}\label{eq:att-view-semantic-attention}
{{\beta }_{f}}=\frac{\exp \left({\frac{1}{\left| {{\mathcal{V}}_{f}} \right|}\sum\limits_{{{v}_{i}}\in {{\mathcal{V}}_{f}}}{{{q}^{T}_{f}}\cdot \sigma \left( {W}'\cdot z_{i}^{f}+{b}' \right)}}\right)}{\sum\nolimits_{f\in \mathcal{F}}{\exp \left({\frac{1}{\left| {{\mathcal{V}}_{f}} \right|}\sum\limits_{{{v}_{i}}\in {{\mathcal{V}}_{f}}}{{{q}^{T}_{f}}\cdot \sigma \left( {W}'\cdot z_{i}^{f}+{b}' \right)}}\right)}},
\end{equation}
where $W'$ and $b'$ are learnable parameter matrix and bias vector, $\sigma(\cdot)$ the activation function, $q_f$ the attention vector of $f$-th type of nodes, and $\mathcal{V}_f$ the set of $f$-th type of nodes. After getting the normalized importance coefficient ${\beta_\mathcal{F}}$, the final node representation of the attribute-guided view $Z^t_{attr}$ is given as
\begin{equation}
{{Z}_{attr}^t}=\sum\nolimits_{f\in \mathcal{F}}{{{\beta }_{f}}\cdot {{Z}^{f}}}.
\end{equation}

\subsection{Encoder for Topology-guided View}
As an important structural property of heterogeneous graphs, meta-paths play an indispensable role in modeling different semantic relationships in heterogeneous graphs. To fully utilize information in graph topology, the topology-guided encoder performs a hierarchical message aggregation process~\cite{han}, including message aggregation within the same meta-path and message aggregation between different meta-paths.

\subsubsection{Message Aggregation within the Same Meta-path} Given a meta-path $\mathcal{M}^p$, many node pairs may be connected by $\mathcal{M}^p$. To distinguish the importance of different meta-path-based neighbors, for the target node $v_i$ and its meta-path $\mathcal{M}^p$ based neighbor nodes $N^{\mathcal{M}^p}_i$, we also adopt a message aggregation process based on the attention mechanism. The weight coefficient between nodes $v_i$ and $v_j$ on $\mathcal{M}^p$ is formalized as
\begin{equation}\label{eq:nodeWeight}
\gamma _{ij}^{{{\mathcal{M}}^{p}}}=\frac{\exp \left( \sigma \left( \text{s}_{{{\mathcal{M}}^{p}}}^{T}\cdot [{{h}_{i}}||{{h}_{j}}] \right) \right)}{\sum\nolimits_{k\in N_{i}^{{{\mathcal{M}}^{q}}}}{\exp \left( \sigma \left( \text{s}_{{{\mathcal{M}}^{p}}}^{T}\cdot [{{h}_{i}}||{{h}_{k}}] \right) \right)}},
\end{equation}
where $\sigma$ is the activation function, $s_{{\mathcal{M}}^{p}}$ the learnable parameter vector specific to meta-path ${\mathcal{M}}^{p}$, and $||$ the concatenate operation. The representation of target node $v_i$ under meta-path $\mathcal{M}^q$ can be aggregated by the neighbor’s representations with the corresponding coefficients as follows
\begin{equation}\label{eq:intraGroupAgg}
h_{i}^{{{\mathcal{M}}^{p}}}=\sigma \left( \sum\nolimits_{j\in N_{i}^{{{\mathcal{M}}^{p}}}}{\gamma _{ij}^{{{M}^{p}}}{{h}_{j}}} \right).
\end{equation}

\subsubsection{Message Aggregation between Different Meta-paths} In a heterogeneous graph, there normally exist multiple meta-paths $\{\mathcal{M}^1, \mathcal{M}^2, ..., \mathcal{M}^P\}$. Accordingly, we can obtain $P$ groups of representations $\{H^{\mathcal{M}^1},H^{\mathcal{M}^2},...,H^{\mathcal{M}^p}\}$ by message aggregation within the same meta-path. We then also use a parameterized attention vector to combine semantic messages from different meta-paths to obtain the final representations $Z_{topo}$. The weighted aggregation between $\{\mathcal{M}^1, \mathcal{M}^2, ..., \mathcal{M}^P\}$ can then be formulated as
\begin{equation}
{{Z}_{topo}}=\sum\nolimits_{p=1}^{P}{{{\eta }^{{{\mathcal{M}}^{p}}}}\cdot {{H}^{{{\mathcal{M}}^{p}}}}}.
\end{equation}
where $\eta^{\mathcal{M}^p}$ is the weight coefficient of meta-path $\mathcal{M}^p$, which can be computed in the same way as Eq.(\ref{eq:att-view-semantic-attention}).

\subsection{Reciprocal Contrastive Optimization} After having the two view encoders on respective guidance of attributes and topology, we can compute node representations from these two views. To learn the final node representation, we introduce a reciprocal contrastive mechanism to integrate and enhance the two views, which includes the definition of positive/negative samples and loss function for the contrastive optimization of the model.

\subsubsection{The Definition of Positive and Negative Sample Pairs} For computing contrastive loss, we define the positive and negative samples in a heterogeneous graph. We first use the representations of the same node under the above two views as a pair of positive samples like the existing methods~\cite{grace}. To improve self-supervised learning, we consider attribute similarity and topological correlation at the same time to expand positive samples, while maintaining sample quality. We calculate the attribute similarity (Eq.(\ref{sim})) of the target type of nodes, and choose each node pair $<v_i , v_j>$ whose attribute similarity $s_{ij}$ is greater than a preset hyper-parameter threshold $\epsilon_{a}$ as the positive sample candidates. We then further determine the positive sample pairs by computing the correlation between the target node and neighbors based on the meta-path. Since different meta-paths in a heterogeneous graph often play different roles, the correlation is determined by the number of meta-paths between the target node and its neighbors on the meta-paths and further assign importance coefficients to meta-paths.

To be specific, given a target node $v_i$ and a meta-path-based node $v_j$, the correlation function ${\mathbb{T}}_i(\cdot)$ is defined as
\begin{equation}
{{\mathbb{T}}_{i}}(v_j)=\sum\nolimits_{p=1}^{P}{{{\delta }^{{{\mathcal{M}}^{p}}}}\cdot {\mathbbm{1}}\left( {{v}_{j}}\in N_{i}^{{{\mathcal{M}}^{p}}} \right)},
\end{equation}
where ${\mathbbm{1}}(\cdot)$ is the indicator function with the value of $0$ or $1$, ${\delta }_{\mathcal{M}^p}$ the importance coefficient of meta-path ${\mathcal{M}^p}$, and ${N^{\mathcal{M}^p}_i}$ the set of neighbors of node $v_i$ on meta-path ${\mathcal{M}^p}$. We choose a node pair $<v_i , v_j>$ whose topological correlation ${\mathbb{T}}_i(v_j)$ is greater than the preset threshold $\epsilon_{t}$ as the positive sample candidates. Finally, the overall positive sample set  of the target node $v_i$ needs to satisfy both attribute similarity and topological correlation, defined as follows
\begin{equation}\label{eq:pp}
\mathbb{P}(i)=\{{{v}_{j}}\ |\ s_{ij}\ge \epsilon_{a} \cap  {{\mathbb{T}}_{i}}({{v}_{j}})\ge \epsilon_{t},\ i\ne j\}.
\end{equation}
Similarly, the overall negative sample set of $v_i$ is
\begin{equation}
\mathbb{N}(i)=\{{{v}_{j}}\ |\ s_{ij}< \epsilon_{a} \cup {{\mathbb{T}}_{i}}({{v}_{j}})< \epsilon_{t},\ i\ne j\}.
\end{equation}

\subsubsection{Contrastive Loss} With the positive sample set $\mathbb{P}(i)$ and negative sample set $\mathbb{N}(i)$ obtained above, we optimize the model by maximizing the agreement between the representations of the positive sample nodes~\cite{grace}. The contrastive loss function for each view can be formulated as
\begin{equation}\label{eq:contrastiveInOneView}
\psi (Z,{Z}')=\frac{1}{|\mathcal{V}|}\sum\limits_{{{v}_{i}}\in \mathcal{V}}{-\log \frac{\sum\limits_{{{v}_{j}}\in \mathbb{P}(i)}{{{e}^{\left\langle {{z}_{i}},{{z}_{j}} \right\rangle }}}+\sum\limits_{{{v}_{j}}\in \mathbb{P}(i)\cup {{v}_{i}}}{{{e}^{\left\langle {{z}_{i}},z_j' \right\rangle }}}}{\sum\nolimits_{{{v}_{j}}\in \mathbb{P}(i)\cup \mathbb{N}(i)}{{{e}^{\left\langle {{z}_{i}},{{z}_{j}} \right\rangle }}+{{e}^{\left\langle {{z}_{i}},z_j' \right\rangle }}}}},
\end{equation}
where $\left\langle {{z}_{i}},{{z}_{i}'} \right\rangle =\cos ({{z}_{i}},{{z}_{i}'})/\tau$ and $\tau$ a temperature parameter. The overall loss function is then defined as the weighted average of the losses of two views, formally given by 
\begin{equation}\label{eq:finalLoss}
{{\mathcal{L}}_{final}}=\lambda\cdot {\psi ({Z_{topo}},{Z_{attr}})}+(1-\lambda)\cdot{\psi ({Z_{attr}},{Z_{topo}})},
\end{equation}
where $\lambda$ is a weighted coefficient to balance these two parts. To this end, we use the concatenation of $Z_{attr}$ and $Z_{topo}$ to perform downstream tasks. The overall process of HGCL is shown in Algorithm 1.

\begin{algorithm}
	\renewcommand{\algorithmicrequire}{\textbf{Input:}}
	\renewcommand{\algorithmicensure}{\textbf{Output:}}
	\caption{The overall process of HGCL}
	
	\label{alg:1}
	\begin{algorithmic}[1]
		\REQUIRE~~\\
		Heterogeneous graph $\mathcal{G}=(\mathcal{V}, \mathcal{E}, \mathcal{X})$; node type set $\mathcal{F}$; meta-path set $\mathcal{M}$; weighted coefficient $\lambda$
		
		\FOR{epoch $\gets 1,2,3...$}
		\STATE $//$encoder for attribute-guided view 
		\FOR {each node type $f \in \mathcal{F}$}
		\STATE  Compute similarity matrix $S^f$ with Eq.(1), obtain the type-specific homogeneous graph $G^f$
		\STATE  Obtain node homogeneous representations $H^f$ of $G^f$ using Eq.(2)
		\ENDFOR

		\FOR{target node type $t$ and other node types $f$ ($t, f \in \mathcal{F}, f\neq t$)}
		\STATE  Regenerate heterogeneous edges with Eq.(3) using $H^t$ and $H^f$
		\STATE  Aggregate heterogeneous neighbor information to obtain representations of target type nodes $H^{t,f}$ with Eq.(6)
		\ENDFOR
		\STATE Combine $H^{t}$ with all $H^{t,f}$ to obtain final representations of the attribute-guided view $Z^t_{attr}$ with Eq.(8)
		\\
		\STATE $//$encoder for topology-guided view 
		\FOR {each meta-path $p\in \mathcal{M}$}
		\STATE  Obtain node representations $H^{\mathcal{M}^p}$ based on the same meta-path with Eq.(10)
		\ENDFOR
		
		\STATE Fuse all $H^{\mathcal{M}}$ from different metapaths to obtain final node representations of the topology-guided view $Z^t_{topo}$ with Eq.(11)

		\STATE $//$contrastive optimization
		
		\STATE Compute positive and negative sample set $\mathbb{P}$ and $\mathbb{N}$ with Eqs.(13) and (14)
		\STATE Compute contrastive loss ${\mathcal{L}}_{final}$ with Eq.(16)
		\STATE Update parameters by applying stochastic gradient ascent to maximize ${\mathcal{L}}_{final}$
		\ENDFOR

		\ENSURE~~\\
		\STATE Final node representations matrix $Z=Z^t_{attr}||Z^t_{topo}$, for use in a downstream task
	\end{algorithmic}
\end{algorithm}
\begin{table}[t]
	\centering
	\caption{Statistics of datasets}\label{tab:datasets}
	\footnotesize
	\resizebox{0.65\linewidth}{!}{
		\begin{tabular}{|c|l|l|l|l|l|}
			\hline
			Datasets & \multicolumn{1}{c|}{Nodes}  &\multicolumn{1}{c|}{Edges}  &\multicolumn{1}{c|}{Meta-paths} &\multicolumn{1}{c|}{Density}               \\ \hline
			ACM     & \begin{tabular}[c]{@{}l@{}}Paper(P):4019\\ Author(A):7167\\ Subject(S):60\end{tabular} & \begin{tabular}[c]{@{}l@{}}P-P:9615\\ P-A:13407\\ P-S:4019\end{tabular}   & \begin{tabular}[c]{@{}l@{}}PAP\\ PSP\end{tabular}       &0.00021    \\ \hline
			Yelp    & \begin{tabular}[c]{@{}l@{}}Business(B):2614\\ User(U):1286\\ Service(S):4\\ Level(L):9\end{tabular}  & \begin{tabular}[c]{@{}l@{}}B-U:30838\\ B-S:2614\\ B-L:2614\end{tabular}      & \begin{tabular}[c]{@{}l@{}}BUB\\ BSB\\ BLB\end{tabular} &0.00235    \\ \hline
			DBLP    & \begin{tabular}[c]{@{}l@{}}Author(A):4057\\ Paper(P):14328\\ Term(T):8789\\ Venue(V):20\end{tabular} & \begin{tabular}[c]{@{}l@{}}A-P:19645\\ P-T:85810\\ P-V:14328\end{tabular} & \begin{tabular}[c]{@{}l@{}}APA\\ APTPA\\ APVPA\end{tabular} &0.00016\\ \hline
			AMiner    & \begin{tabular}[c]{@{}l@{}}Paper(P):20201\\ Author(A):8052\end{tabular} & \begin{tabular}[c]{@{}l@{}}P-P:64671\\P-A:32028 \end{tabular} & \begin{tabular}[c]{@{}l@{}}PPP\\ PAP\end{tabular} &0.00012\\ \hline
		\end{tabular}
	}
\end{table}


\subsection{Time Complexity Analysis}
The newly proposed HGCL contains three components: the attribute-guided view, the topology-guided view, and the reciprocal contrastive optimization module. The computational overhead in attribute-guided views includes the generation of homogeneous/heterogeneous edges and message aggregation between nodes of the same/different types, where the generation of homogeneous edges is preprocessed. The original time complexity of heterogeneous edge generation based on cosine similarity is O$(N_{t}N_{o})$, where $N_{t}$ is the number of target type nodes, and $N_{o}$ is the number of nodes with the largest number of other types of nodes. Here, we further use the KD-tree quick algorithm~\cite{kd-tree} to reduce the complexity of cosine similarity to O$(log_{2}N_{t})$. The message passing mechanism between nodes of the same/different types is the common single-layer ConvGNN, and the time complexity is O$(|E|)$, where $|E|$ is the number of edges based on the meta-path. The topology-guided view adopts a hierarchical message passing process based on the attention mechanism, and the time complexity is O$(N_{t})$+O$(|E|)$. The selection of positive and negative sample pairs in the reciprocal contrastive optimization module is preprocessed, and the loss function adopts the classic InfoNCE-based loss with a time complexity of O$(N_{t}^2)$. In summary, the complete time complexity of our HGCL approach is: O$(log_{2}N_{t} +|E|+|E|+N_{t}+|E|+N_{t}^2)$= O$(|E|+N_{t}^2)$. The time complexity of most ConvGNNs is O$(|E|)$, and the time complexity of InfoNCE-based contrastive loss is O$(N_{t}^2)$. Therefore, our method does not significantly increase the time complexity and can be scalable to larger datasets when scalable InfoNCE-based self-supervised GNNs is used.

\section{Experiments}\label{shiyan}
We first discuss the experimental setup, including datasets, baseline methods used, and detailed experimental settings. We then present the comparison results on node classification and node clustering. We discuss additional experiments to demonstrate the robustness and generality of the new HGCL approach and carry out a parameter analysis.

\subsection{Experimental Setup}

\paragraph{Datasets}
To analyze the effectiveness of HGCL, we performed a comprehensive experimental analysis using four widely-used heterogeneous graph datasets (Table~\ref{tab:datasets}). 
\begin{enumerate}
	\item ACM\footnote{http://dl.acm.org/}~\cite{han}: We extracted a subset of ACM for 4019 papers (P), 7167 authors (A), and 60 subjects (S). We conducted experimental analysis on paper nodes in the ACM dataset. The papers were labeled according to their fields.
	\item Yelp\footnote{https://www.yelp.com/dataset}~\cite{yelp}: We extracted a subset from Yelp Open Dataset containing 2614 businesses (B), 1286 users (U), 4 services (S), and 9 rating levels (L). We conducted experimental analysis on business nodes in the Yelp dataset. The business nodes were labeled by their categories. 
	\item DBLP\textsuperscript{\ref{dblp}}~\cite{dblp}: We extracted a subset of DBLP containing information of 4057 authors (A), 14328 papers (P), 8789 terms (T), and 20 venues (V). We conducted experimental analysis on author nodes in the DBLP dataset. The authors were divided into four research areas. 
	\item AMiner\footnote{https://www.aminer.cn/data/}~\cite{aminer}: We extracted a subset from AMiner containing information of 20201 papers (P) and 8052 authors (A). We conducted experimental analysis on Paper nodes in the AMiner dataset. The papers were labeled according to their fields.
\end{enumerate}

\paragraph{Baseline Methods}
We compared the new HGCL approach with thirteen state-of-the-art embedding methods. These methods include six semi-supervised methods (GraphSAGE~\cite{sage}, GAT~\cite{gat}, HAN~\cite{han}, MAGNN~\cite{magnn}, HGSL~\cite{hgsl}, and Simple-HGN~\cite{DBLP:conf/kdd/LvDLCFHZJDT21}) and seven unsupervised methods (Mp2vec~\cite{metapath2vec}, DGI~\cite{dgi}, GMI~\cite{gmi}, DMGI~\cite{dmgi}, GAE~\cite{gae}, PT-HGNN~\cite{DBLP:conf/kdd/JiangJFSLW21}, and HeCo~\cite{heco}). They can be grouped into four methods for homogeneous graphs (GAT, DGI, GMI, and GAE) and eight methods for heterogeneous graphs (HAN, MAGNN, HGSL, Simple-HGN, PT-HGNN, Mp2vec, DMGI, and HeCo).

\paragraph{Detailed Settings}
For a fair comparison of all methods, the final embedding dimensions evaluated were set to 64. For semi-supervised methods (GAT, HAN, and MAGNN), the labeled nodes were divided into training, validation, and testing sets in the ratio of 10\%, 10\%, and 80\% as done in existing works. For homogeneous methods (GAT, DGI, GMI, and GAE), we tested all their meta-paths and report here the best performance of all methods compared.

For our HGCL approach, we use cross-validation to set the parameters and adopt the Adam optimizer with a learning rate of 0.0001. We set the temperature parameter  $\tau$ to 0.4, the topological correlation threshold for sample selection $\epsilon_{t}$ to 1.0, and the loss weighted coefficient $\lambda$ to 0.5. We set the dimension of the hidden layer to 128. We search the attribute similarity threshold $\epsilon_{a}$ in \{0.5, 0.6, 0.7\} for each dataset, and the weight coefficient for meta-path $\delta$ from 0.0 to 1.0 with a step size of 0.2.

\begin{table*}[t]
	\caption{Performance evaluation of node classification on four datasets (Macro-F1)}\label{tab:classification}
	\large
	\resizebox{\linewidth}{!}{
		\begin{tabular}{ccccccccccccccc}            
			\midrule[1.4pt]
			\multirow{2}{*}{Datasets}  & \multirow{2}{*}{Training} & \multicolumn{6}{c}{Semi-supervised} & \multicolumn{5}{c}{Unsupervised}                       \\ \cmidrule(r){3-8} \cmidrule(r){9-14}
			 &   & \revise{GraphSAGE} & GAT     & HAN     & MAGNN     &HGSL   &Simple-HGN   & Mp2vec & DGI    & DMGI       &PT-HGNN   & HeCo        & HGCL            \\ \midrule
			\multirow{4}{*}{ACM}      & 20\%          &\revise{0.8611}            &0.8966  &0.9062  &0.8693    &0.9131  &0.9203     &0.7011       &0.9041 &\textbf{0.9222} &0.9188   &0.8637    &\ul{0.9206} \\
			                          & 40\%            &\revise{0.8635}          &0.8975  &0.9102  &0.8889    &0.9188  &0.9235     &0.7043       &0.9042 &\ul{0.9251} &0.9201   &0.8784   &\textbf{0.9259} \\
			                         & 60\%         &\revise{0.8688}             &0.8993  &0.9128  &0.8985    &0.9229  &0.9267     &0.7073       &0.9062 &\ul{0.9278} &0.9223   &0.8863   &\textbf{0.9301} \\
			                         & 80\%           &\revise{0.8697}           & 0.8960  & 0.9150  & 0.9064    &0.9276  &\ul{0.9289}& 0.7113       & 0.9055  &0.9256 &0.9219   &0.8937    &\textbf{0.9303} \\ \midrule 
			\multirow{4}{*}{Yelp}        & 20\%           &\revise{0.6254}           & 0.5407  & 0.7724  & 0.8676    &0.9097  &\ul{0.9113}     &0.5396  & 0.5407        & 0.7273 &0.8642    &0.5395        &\textbf{0.9137} \\
			                            & 40\%            &\revise{0.6255}          & 0.5407  & 0.7848  &0.8871    &0.9126  &\ul{0.9225}  & 0.5400  & 0.5407         & 0.7381 &0.8689    &0.5399        &\textbf{0.9263} \\
			                           & 60\%         &\revise{0.6251}             & 0.5400  & 0.7858  &0.9018    &0.9187  &\ul{0.9289}  & 0.5396  & 0.5400         & 0.7448 &0.8714    & 0.5396       &\textbf{0.9324} \\
			                           & 80\%          &\revise{0.6227}            & 0.5381  & 0.7893  &0.8991    &0.9219 &\ul{0.9287}  & 0.5370  & 0.5381         & 0.7541 &0.8705     & 0.5372       &\textbf{0.9294}\\ \midrule
			\multirow{4}{*}{DBLP}     & 20\%      &\revise{0.8872}                & 0.9040  & 0.9221  & \ul{ 0.9381}    &0.9306  &0.9311  & 0.7666  & 0.8851          & 0.9290 &0.9188      & 0.9041        &\textbf{0.9465}\\
			                          & 40\%      &\revise{0.8881}                & 0.9061  & 0.9244  & \ul{ 0.9391}    &0.9327  &0.9339  & 0.8214  & 0.8849         & 0.9296 &0.9203      & 0.9111        &\textbf{0.9471} \\
			                           & 60\%     &\revise{0.8895}                 & 0.9073  & 0.9251  & \ul{ 0.9394}    &0.9342  &0.9375  & 0.8425  & 0. 8861        & 0.9317 &0.9241     & 0.9158        &\textbf{0.9472}\\
			                           & 80\%       &\revise{0.8902}               & 0.9088   & 0.9271  & \ul{ 0.9417}    &0.9388  &0.9412  & 0.8420  & 0.8856        & 0.9333 &0.9269      & 0.9151        &\textbf{0.9479} \\ \midrule 
			\multirow{4}{*}{AMiner}     & 20\%  &\revise{0.8744} &0.8941   &0.8879  &0.8959 &0.8643 &0.9008  &0.6387  &0.8482   &0.8709 &\textbf{0.9108}  &0.9024 &\ul{0.9104}  \\
			                          & 40\%   &\revise{0.8750}                   &0.8950   &0.8895  &0.8988 &0.8658 &0.9053   &0.6436   &0.8521   &0.8743 &\ul{0.9127}  &0.9046   &\textbf{0.9150}  \\
    			                           & 60\%     &\revise{0.8767}                &0.8958   &0.8898  &0.9007 &0.8715 &0.9079   &0.6464   &0.8547   &0.8760 &\ul{0.9149}   &0.9065  &\textbf{0.9173} \\
			                           & 80\%         &\revise{0.8782}            &0.8949   &0.8887  &0.9011 &0.8715 &0.9065  &0.6449   &0.8545   &0.8749 &\ul{0.9146}   &0.9055  &\textbf{0.9171}  \\ \midrule 
		\end{tabular}
	}
\end{table*}

\subsection{Node Classification}
Node classification is a traditional task for the evaluation of the quality of learned node representations. After learning node representations, we adopted a linear support vector machine (SVM)~\cite{svm} classifier to classify nodes. Because the labels for the training and validation sets have been used in semi-supervised methods, to make a fair comparison, we only classified the nodes in the test set in each dataset. We fed node representations to the SVM classifier with varying training ratios from 20\% to 80\%. We repeated experiments 10 times and report here the Macro-F1 and Micro-F1, which are commonly used evaluation metrics.

The results are shown in Table~\ref{tab:classification} and Table~\ref{tab:classification2}, where the best results are in bold fonts and the second-best results are underlined. The new HGCL approach outperformed all baseline methods compared under different training ratios, including six semi-supervised methods and five unsupervised methods, in most cases considered. Among them, PT-HGNN and HeCo are the most advanced self-supervised methods based on contrastive learning, but they highly rely on the trustworthiness of the initial graph topology. In particular, the proposed HGCL outperformed HeCo by 11.96\% in Macro-F1 and 6.90\% in Micro-F1 on average. The superior performance of the new approach may be mainly due to the design of both attribute- and topology-guided views to maximize the information in attributes and topology, respectively, and the use of the high-quality sample selection mechanism to achieve mutual supervision and mutual enhancement of the two views, making it robust to noisy graphs. All GNN-based baseline methods (including HeCo) employ a single fusion mechanism in which attributes are propagated and aggregated under the guidance of initial topology, causing noise in and disparity of attributes and topology to degrade model performance.

\begin{table*}[t]
	\caption{Performance evaluation of node classification on four datasets (Micro-F1)}\label{tab:classification2}
	\large 
	\resizebox{\linewidth}{!}{
		\begin{tabular}{ccccccccccccccc}            
			\midrule[1.4pt]
			\multirow{2}{*}{Datasets}  & \multirow{2}{*}{Training} & \multicolumn{6}{c}{Semi-supervised} & \multicolumn{5}{c}{Unsupervised}                       \\ \cmidrule(r){3-8} \cmidrule(r){9-14}
			 &     &\revise{GraphSAGE} & GAT     & HAN     & MAGNN     &HGSL   &Simple-HGN   & Mp2vec & DGI    & DMGI       &PT-HGNN   & HeCo        & HGCL            \\ \midrule
			\multirow{4}{*}{ACM}       & 20\%         &\revise{0.8607}             & 0.8955  & 0.9056  & 0.8703    &0.9147  &0.9170         & 0.7444       & 0.9033  &\textbf{0.9207} &0.9152 &0.8696   &\ul{0.9196}\\
			                           & 40\%       &\revise{0.8623}               & 0.8968  & 0.9099  & 0.8895    &0.9182  &0.9204        & 0.7480       & 0.9034  &\ul{0.9236} &0.9179   &0.8807     &\textbf{0.9248}\\
			                           & 60\%    &\revise{0.8623}                  & 0.8984  & 0.9123  & 0.8985    &0.9213 &0.9239          & 0.7522    & 0.9051     &\ul{0.9260} &0.9211   &0.8875     &\textbf{0.9290}\\
			                           & 80\%       &\revise{0.8651}               & 0.8950  & 0.9142  & 0.9061    &0.9255  &\ul{0.9262}          & 0.7557  & 0.9044       &\ul{0.9238} &0.9207  &0.8949    &\textbf{0.9295}\\ \midrule 
			\multirow{4}{*}{Yelp}        & 20\%         &\revise{0.7677}             & 0.7306  & 0.7885  &0.8711    &0.9056  &\textbf{0.9091}    & 0.7289  & 0.7306       & 0.7833 &0.8944    & 0.7289    &\ul{0.9084} \\
			                            & 40\%         &\revise{0.7690}             & 0.7314  & 0.7992  &0.8887    &0.9100  &\ul{0.9166}    & 0.7295  & 0.7314      & 0.7893 &0.8980     & 0.7298        &\textbf{0.9197} \\
			                           & 60\%            &\revise{0.7665}          & 0.7296  & 0.7997  &0.9034    &0.9132  &\ul{0.9251}    & 0.7297  & 0.7296       & 0.7934 &0.9046     & 0.7297        &\textbf{0.9269}\\
			                           & 80\%         &\revise{0.7649}             & 0.7281  & 0.8041  &0.9008    &0.9187  &\ul{0.9233}    & 0.7278  & 0.7281       & 0.8000 &0.9015     & 0.7280       &\textbf{0.9241} \\ \midrule
			\multirow{4}{*}{DBLP}     & 20\%         &\revise{0.8943}             & 0.9105  & 0.9269  & \ul{ 0.9420}    &0.9299  &0.9355   & 0.7761  & 0.8926        & 0.9339 &0.9214      & 0.9112        &\textbf{0.9502} \\
			                            & 40\%       &\revise{0.8927}               & 0.9126  & 0.9290  & \ul{ 0.9428}    &0.9321  &0.9376   & 0.8289  & 0.8920       & 0.9344 &0.9252      & 0.9179        &\textbf{0.9506} \\
			                           & 60\%          &\revise{0.8943}            & 0.9135  & 0.9300  & \ul{ 0.9432}    &0.9387  &0.9402   & 0.8502  & 0.8934        & 0.9364 &0.9279       & 0.9224        &\textbf{0.9508} \\
			                          & 80\%         &\revise{0.8951}             & 0.9148  & 0.9318  & \ul{ 0.9453}    &0.9419  &0.9431  & 0.8495   & 0.8927        & 0.9378 &0.9311       & 0.9213        &\textbf{0.9512}\\  \midrule 
			\multirow{4}{*}{AMiner}     & 20\%     &\revise{0.8711}        &0.8919   &0.8854  &0.8949 &0.8627 &0.8933   &0.6465   &0.8485   &0.8718 &\ul{0.9101}  &0.9010   &\textbf{0.9102}\\
			                            &40\%       &\revise{0.8718}              &0.8929   &0.8870  &0.8983 &0.8650 &0.8952   &0.6506   &0.8522   &0.8751 &\ul{0.9119}   &0.9032 &\textbf{0.9138}  \\
			                           &60\%       &\revise{0.8930}               &0.8939   &0.8863  &0.8995 &0.8684 &0.8990   &0.6534   &0.8547   &0.8769 &\ul{0.9135}   &0.9052  &\textbf{0.9163}  \\
			                           & 80\%      &\revise{0.8933}                &0.8930   &0.8863  &0.8997 &0.8689 &0.8976   &0.6535   &0.8546   &0.8759 &\ul{0.9133}   &0.9046  &\textbf{0.9163} \\  \midrule 
		\end{tabular}
	}
\end{table*}

\begin{center}
	\begin{table*}[t]
		\centering
		\caption{Performance evaluation of node clustering on four datasets}\label{tab:clustering}
		\scriptsize
		\resizebox{0.99\linewidth}{!}{
			\begin{tabular}{cclcccccccc}
				\midrule[1.05pt]
				Datasets               & \multicolumn{2}{c}{Metrics} & Mp2vec & GAE    & DGI             & GMI             & DMGI   &PT-HGNN        & HeCo            & HGCL           \\ \midrule
				\multirow{2}{*}{ACM}  & \multicolumn{2}{c}{NMI}     & 0.3765   & 0.4059 &0.5183    & 0.3763       & 0.6065 &\ul{0.6477}  & 0.6316   & \textbf{0.6927} \\
				& \multicolumn{2}{c}{ARI}     & 0.3025   & 0.3319 &0.4374    & 0.3022       & 0.5827 &\ul{0.6892}  & 0.6649   & \textbf{0.7377} \\ \midrule
				\multirow{2}{*}{Yelp} & \multicolumn{2}{c}{NMI}     & 0.3890   & 0.3919 & 0.3942  & 0.3942  & 0.3690 &\ul{0.4063}  &0.3942  &\textbf{0.4137}\\
				& \multicolumn{2}{c}{ARI}     & 0.4249   & 0.4257 & 0.4262  & 0.4260       & 0.3346 &\ul{0.4298}  &0.4262 &\textbf{0.4302} \\ \midrule
				\multirow{2}{*}{DBLP} & \multicolumn{2}{c}{NMI}     & 0.5451   & 0.5257 & 0.6637  & 0.4101       & \ul{0.7388} &0.7232 & 0.6942 &\textbf{0.8091} \\
				& \multicolumn{2}{c}{ARI}     & 0.5815   & 0.4986 & 0.6838  & 0.4056       & \ul{0.7929} &0.7815 & 0.7483 &\textbf{0.8554} \\ \midrule
				\multirow{2}{*}{AMiner} & \multicolumn{2}{c}{NMI}   &0.3923  &0.4177   &0.5942    &0.3913    &0.5232 &\textbf{0.6103}  &0.6078  &\ul{0.6082}  \\
				& \multicolumn{2}{c}{ARI}     &0.2861     &0.3009   &0.6130    &0.2925    &0.5081 &\ul{0.6395}  &0.5773  &\textbf{0.6470} \\ \midrule[1.05pt]
			\end{tabular}
		}
	\end{table*}
\end{center}

\subsection{Node Clustering}
Our unsupervised HGCL approach is particularly suitable for this unsupervised task. For comparison, we chose seven unsupervised methods (Mp2vec, GAE, DGI, GMI, DMGI, PT-HGNN and HeCo) as baselines. We excluded semi-supervised methods since they use the labels of some of the training data. We applied the $k$-Means algorithm 10 times to the learned node representations. We report the average normalized mutual information (NMI) and adjusted rand index (ARI) for comparison. 
\begin{figure}[ht]  
	\centering       
	\subfloat[Edge deletion]
	{
		\begin{minipage}[t]{0.4\textwidth}
			\centering          
			\includegraphics[width=1.0\textwidth]{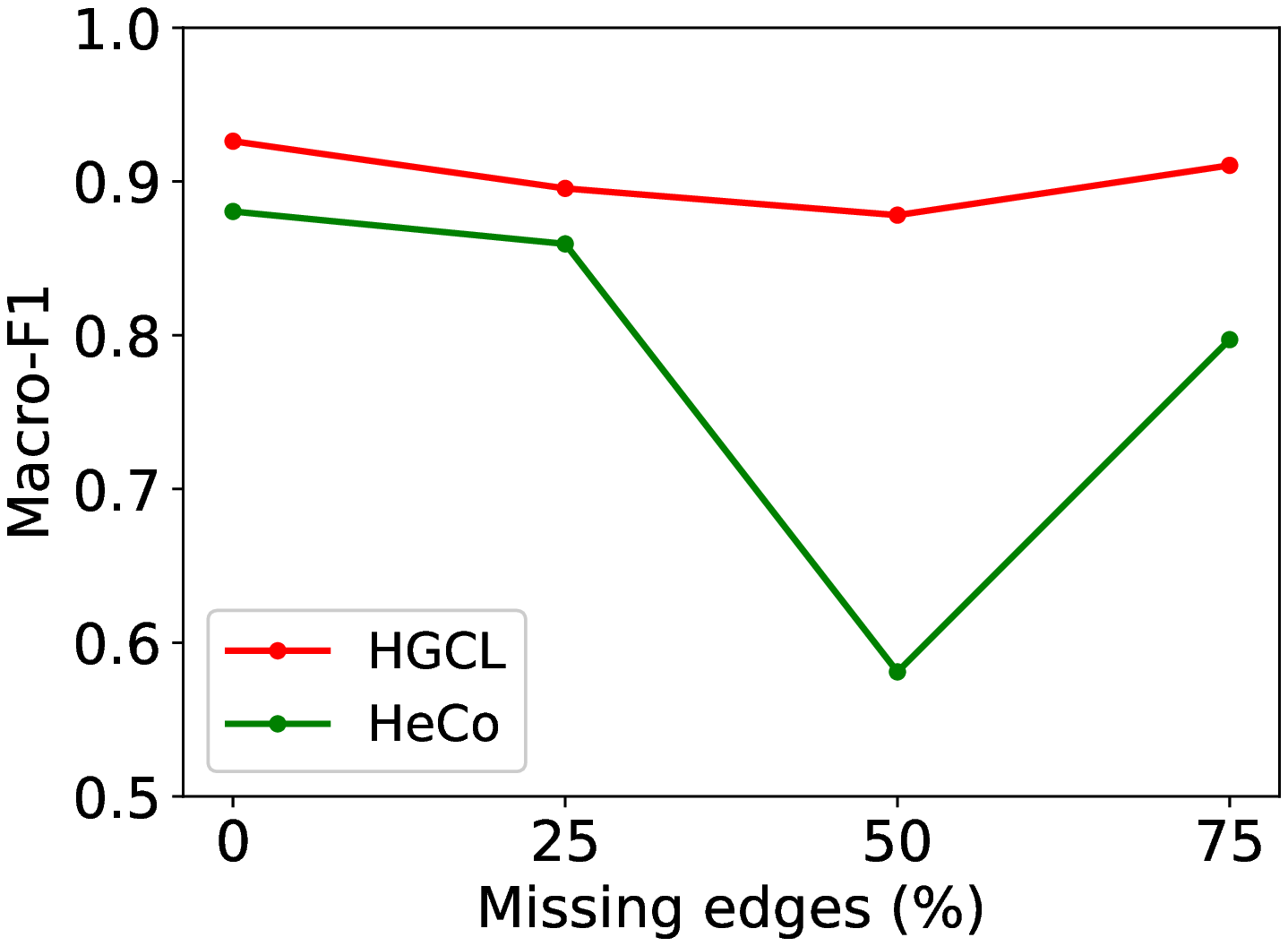}  
		\end{minipage}%
	}
	\subfloat[Attribute masking]
	{
		\begin{minipage}[t]{0.4\textwidth}
			\centering      
			\includegraphics[width=1.0\textwidth]{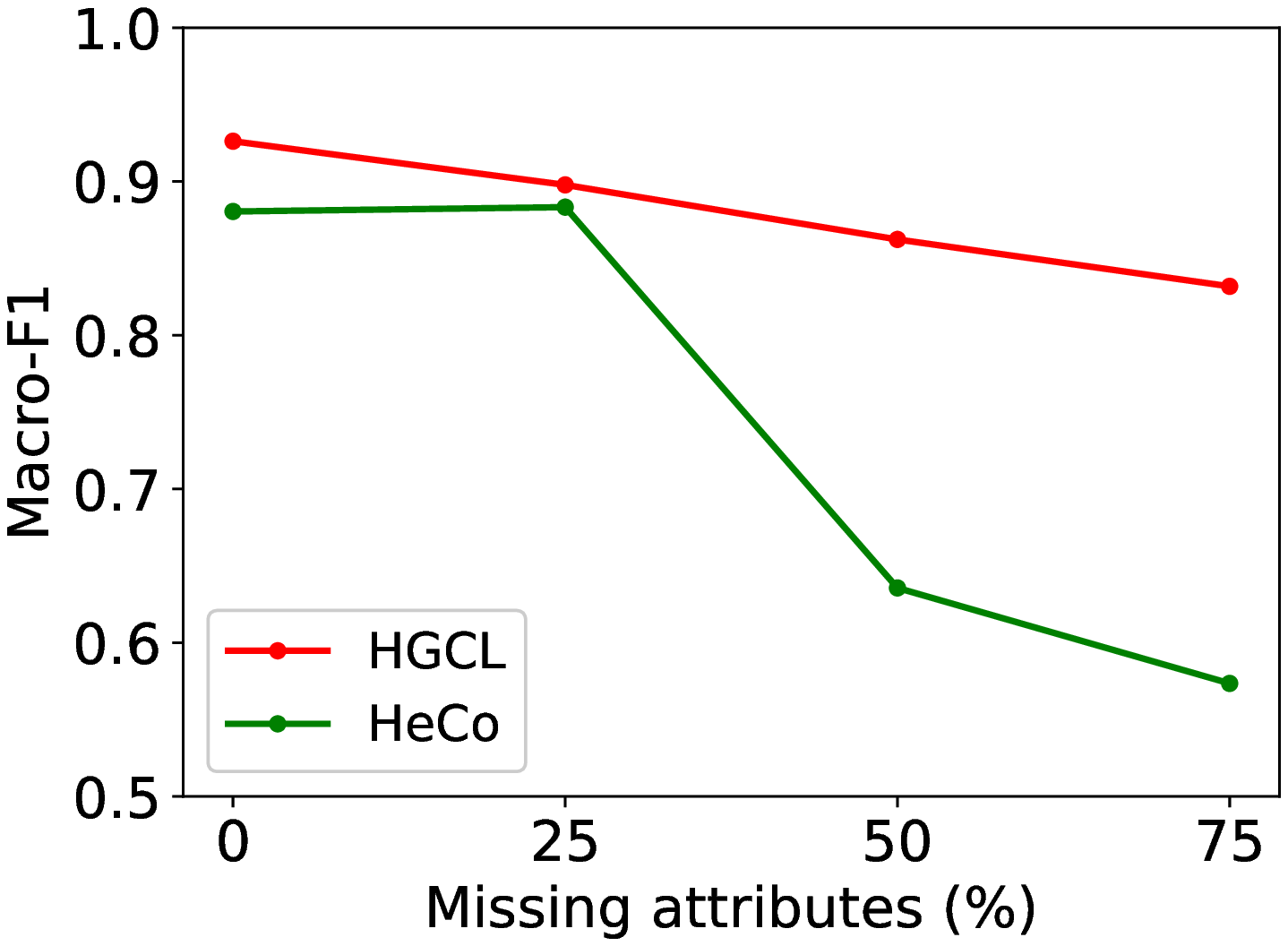}   
		\end{minipage}
	}%
	\caption{The average result of node classification under different training ratios. We report Macro-F1 on the ACM dataset as a case study.}\label{lubang}  
\end{figure}

\begin{figure*}[ht]  
	\centering       
	\subfloat
	{
		\begin{minipage}[t]{0.72\textwidth}
			\centering          
			\includegraphics[width=0.99\textwidth]{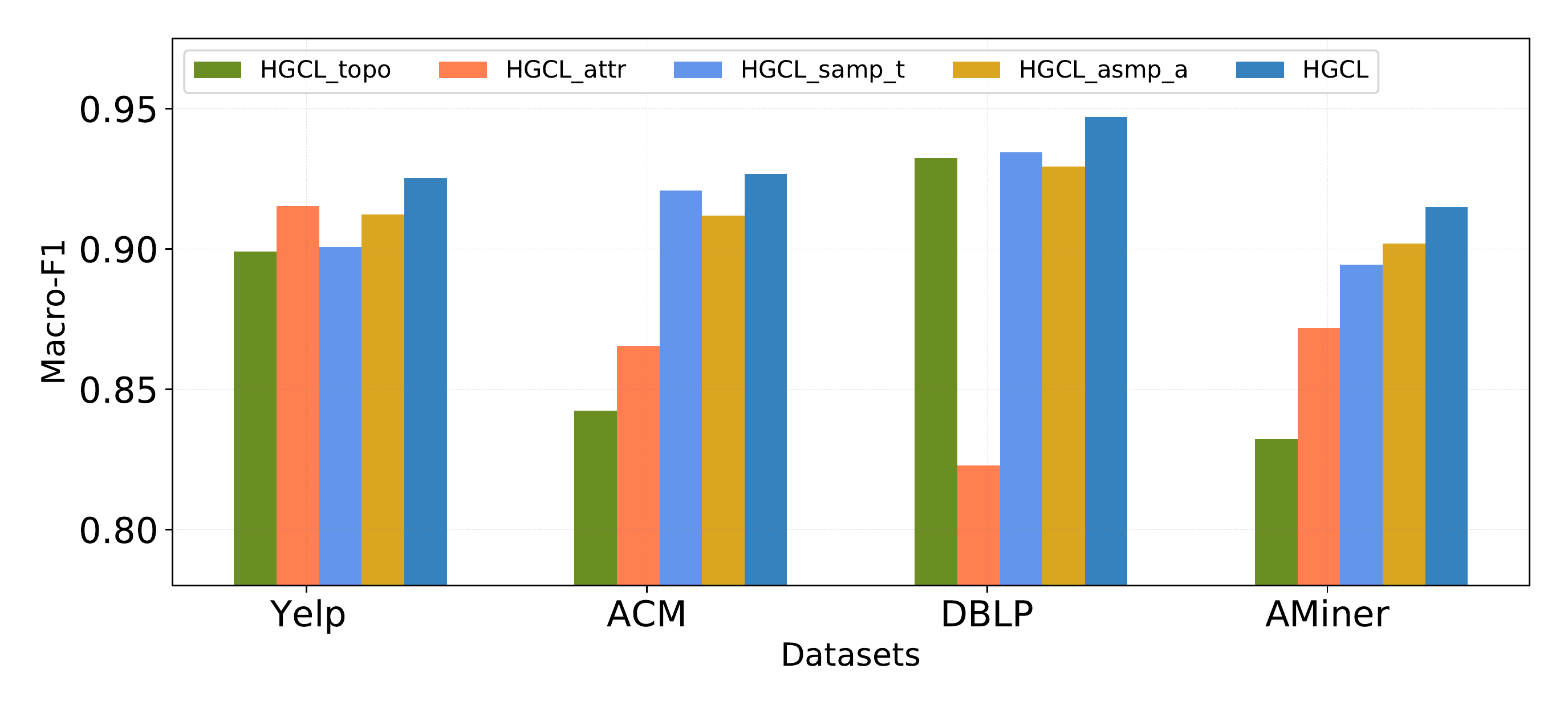}  
		\end{minipage}%
	}
 
	\subfloat
	{
		\begin{minipage}[t]{0.72\textwidth}
			\centering      
			\includegraphics[width=0.99\textwidth]{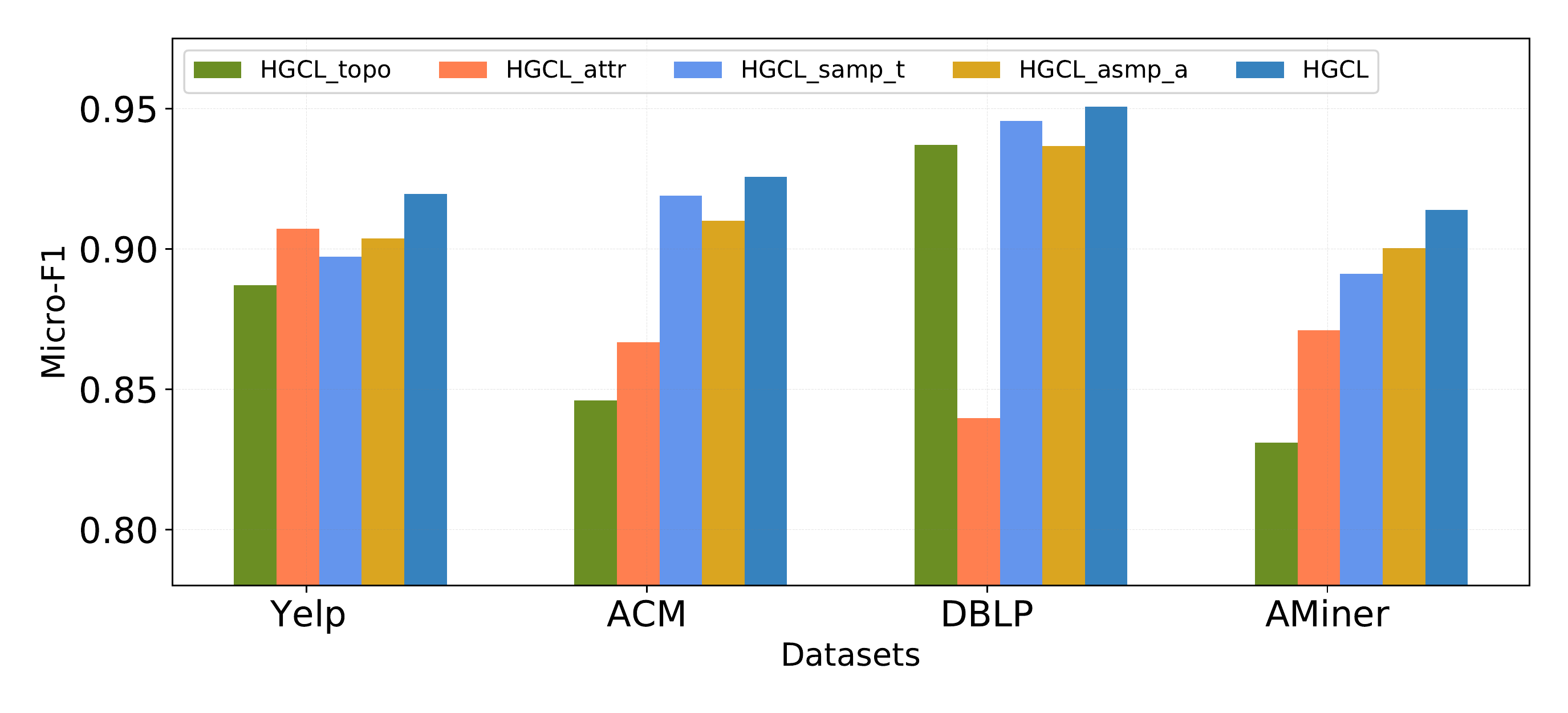}   
		\end{minipage}
	}%
	\caption{The average result of node classification under different training ratios. We report Macro-F1 (in the upper) and Micro-F1 (in the lower) on four datasets as the ablation study.}\label{fig:lubang}  
\end{figure*}
The experimental results are presented in Table~\ref{tab:clustering}. As shown, HGCL outperformed all baseline methods in terms of both NMI and ARI in most cases. For example, the proposeed HGCL outperformed the best baseline methods (DMGI and PT-HGNN) by 3.06\% in NMI and 2.97\% in ARI on average on four datasets. The superior performance of HGCL in node clustering over the state-of-the-art methods further demonstrates the effectiveness of the new method.

\subsection{Robustness Verification}
To evaluate the robustness of HGCL, we constructed original graphs
with random edge deletion or random attribute masking. Specifically, for each pair of nodes in the original graph, we randomly removed an edge (if it exists) with a probability 25\%, 50\%, and 75\%~\cite{dele}. For random attribute masking, we randomly masked a fraction of dimensions with zeros in node initial features and set the masking ratio to 25\%, 50\%, and 75\% for experiments. As shown in Fig.~\ref{lubang}, compared with the baseline HeCo, which designs a contrastive mechanism based solely on pre-defined structural properties in the graph, HGCL achieves better and more stable results in both scenarios. While HeCo has large fluctuations when the edge deletion probability increases, and fails completely with increasing attribute masking ratio, HGCL performs reasonably well. This is because the proposed contrastive mechanism of mining key information in attributes and topology separately and integrating and enhancing them with each other reduce the interference of noise to the model and greatly improves the robustness.  

\subsection{Ablation Study} 
To gain deeper insights into the contributions of different components introduced in our approach, we designed four variants of HGCL, i.e., HGCL\_topo, HGCL\_attr, HGCL\_samp\_t, and  HGCL\_samp\_a. In HGCL\_topo, nodes are encoded only in topology-guided view, and the representations of corresponding positive and negative samples only come from the topology-guided view. Similarly, for HGCL\_attr, nodes are only encoded in the attribute-guided view, and the representations also only come from the attribute-guided view. In HGCL\_samp\_t, we only use topological correlation to select positive and negative samples to guide model training; for HGCL\_samp\_a, we only use attribute similarity to select positive and negative samples. We compared these four variants with the complete HGCL for the task of node classification as an example.

As shown in Fig.~\ref{fig:lubang}, HGCL performed consistently better than its four variants on all four datasets. In addition, HGCL\_attr and HGCL\_topo exhibited different strengths on different datasets, suggesting the need to maximize and integrate their advantages to learn better node representations. Likewise, the strengths of HGCL\_samp\_t and HGCL\_samp\_a are not the same on different datasets, which indicates the necessity of considering both attribute similarity and topological correlation when there is noise in the graph, and further demonstrates our proposed high-quality sample selection mechanism achieves effective supervision for model training.
\begin{figure}[hhhhht]  
	\centering       
	\includegraphics[width=0.60\linewidth]{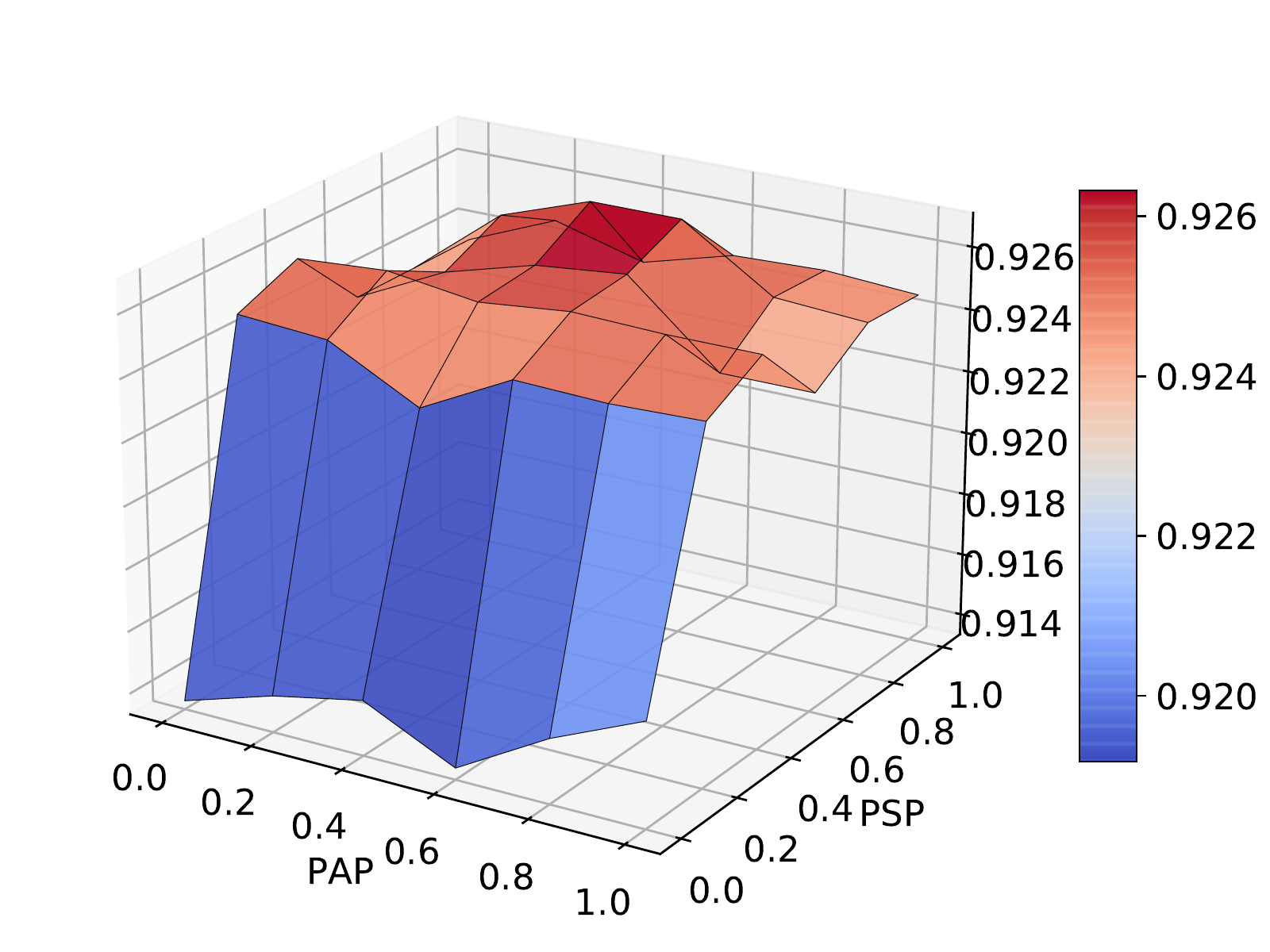} 
	\caption{Parameter sensitivity analysis of $\delta$ on ACM dataset. Shown are the average results of node classification under different training ratios. A warmer color denotes a higher accuracy.}\label{fig:metapath}  
\end{figure}

\begin{figure*}[hhhhht]  
	\centering    
	\subfloat[ACM] 
	{
		\begin{minipage}[t]{0.4\textwidth}
			\centering      
			\includegraphics[width=1.0\textwidth]{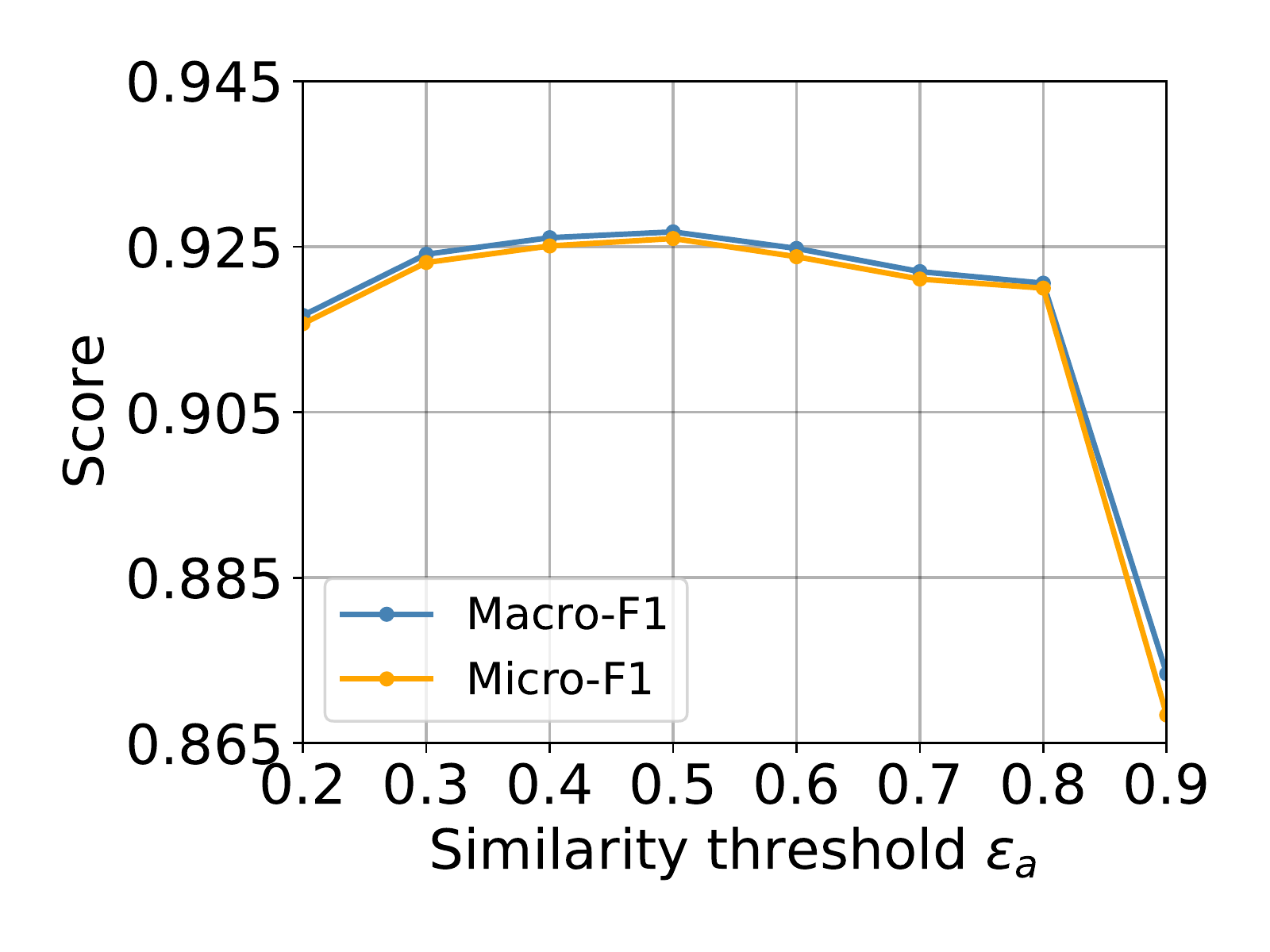}   
		\end{minipage}
	}%
	\subfloat[Yelp] 
	{
		\begin{minipage}[t]{0.4\textwidth}
			\centering          
			\includegraphics[width=1.0\textwidth]{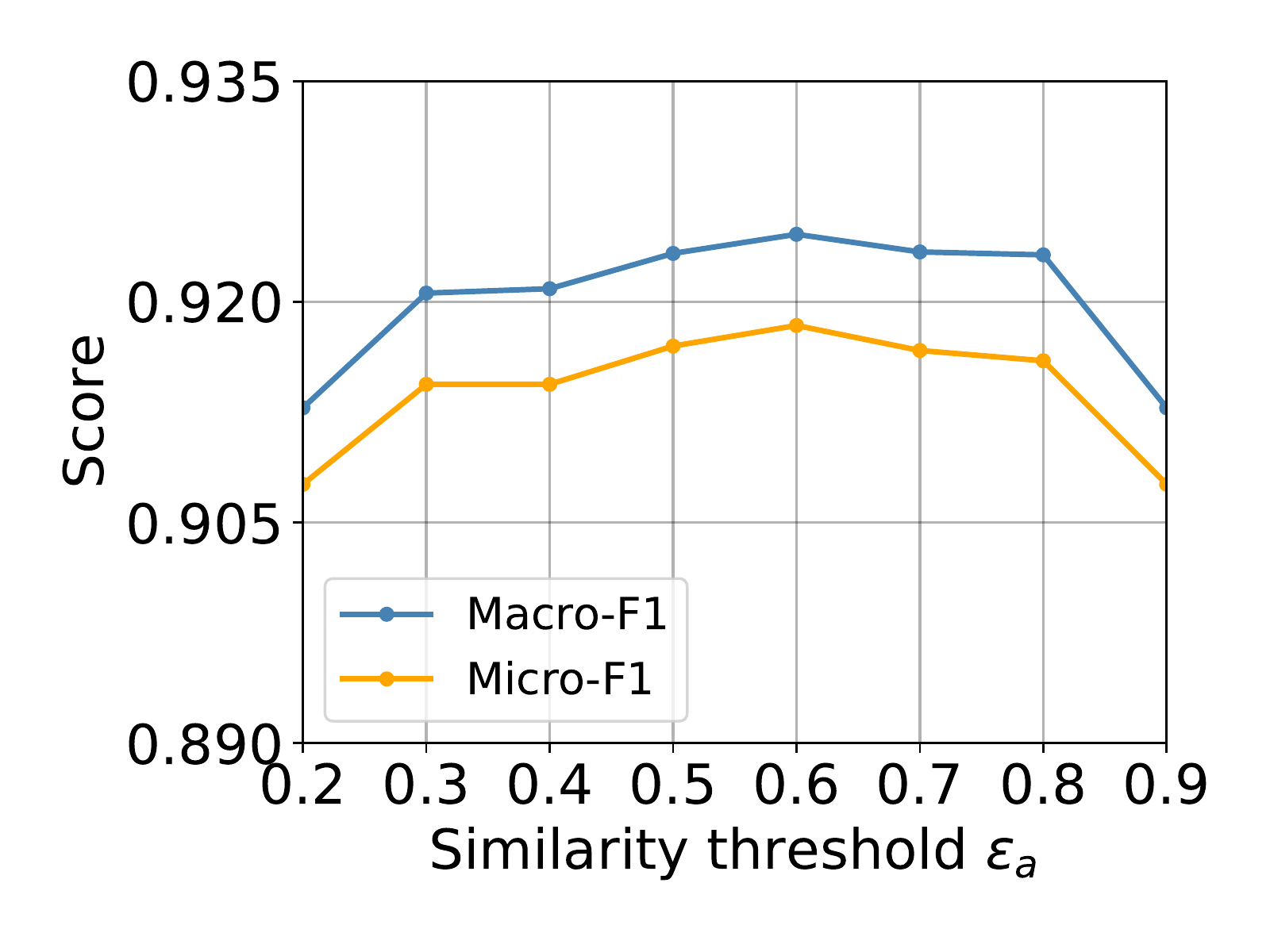}   
		\end{minipage}%
	}
 
	\subfloat[DBLP] 
	{
		\begin{minipage}[t]{0.4\textwidth}
			\centering          
			\includegraphics[width=1.0\textwidth]{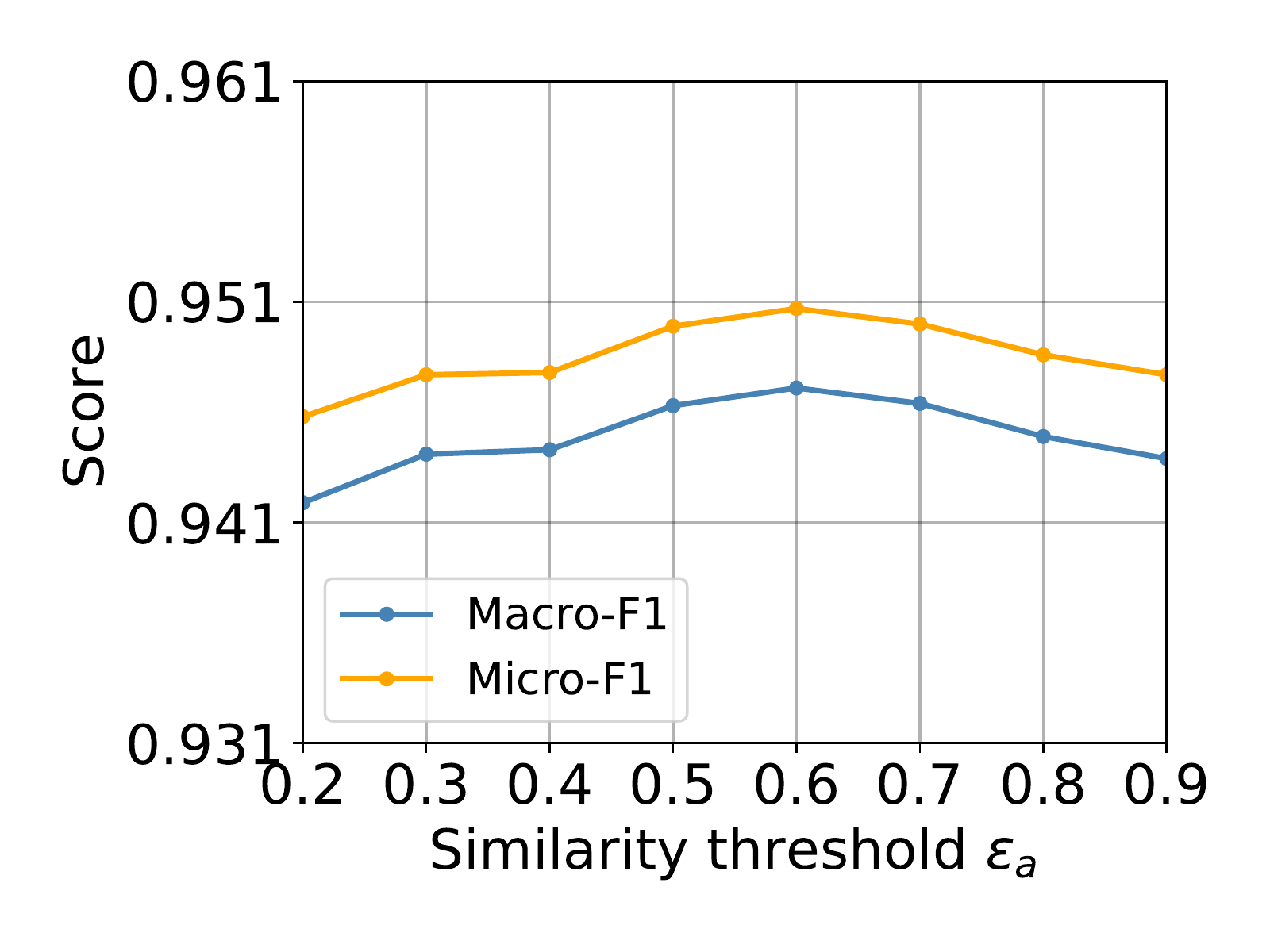}   
		\end{minipage}%
	}
	\subfloat[AMiner] 
	{
		\begin{minipage}[t]{0.4\textwidth}
			\centering          
			\includegraphics[width=1.0\textwidth]{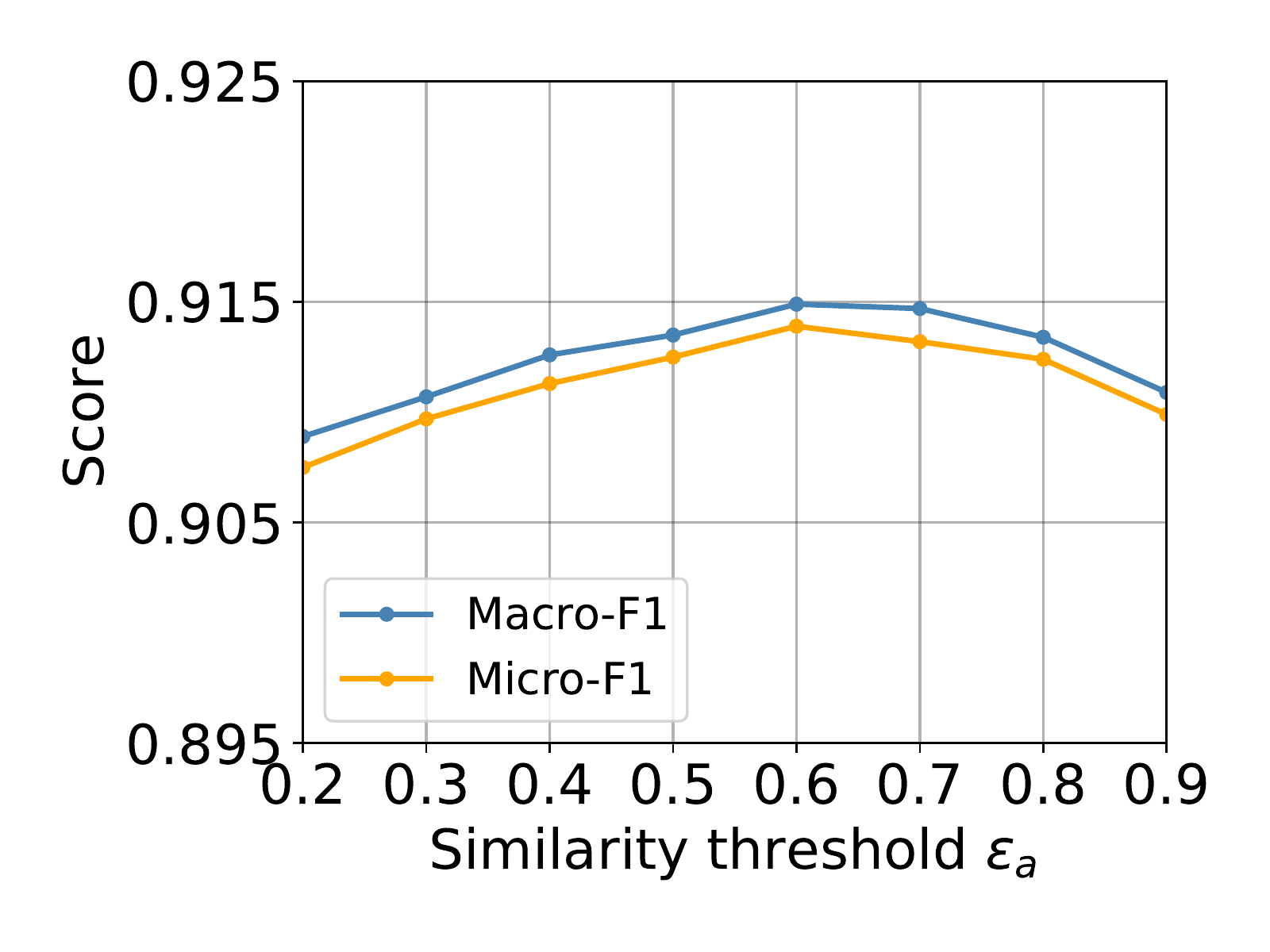}   
		\end{minipage}%
	}
	
	\caption{Parameter sensitivity analysis of $\epsilon$. We report the average result of the node classification task under different training ratios on four datasets.}\label{fig:sim}
\end{figure*}
\subsection{Parameter Analysis}
We investigated the sensitivity of two key hyperparameters: the weighted coefficient $\delta$ of meta-path and the attribute similarity threshold $\epsilon$. We report the average result of node classification with different training ratios on these three datasets.

\textbf{The weighted coefficient $\delta$ of meta-path.} We chose “paper-author-paper” (PAP) and “paper-subject-paper” (PSP) used in the ACM dataset to analyze $\delta$. HGCL achieved a good performance when $0.2\le {{\delta }_{PSP}}\le 0.8$, and was not so sensitive to ${{\delta }_{PAP}}$ (Fig.~\ref{fig:metapath}). This is reasonable since papers that share the same subject tend to belong to the same field, and thus PSP is more important than PAP. Therefore, the well-designed meta-paths can help select high-quality positive samples to further improve the performance of HGCL. Note that we still needed to use a suitable value of ${{\delta }_{PAP}}$, because we set the topological correlation threshold $\epsilon_{t}$ in Eq.(\ref{eq:pp}) as a constant, and ${{\delta }_{PAP}}$ and ${{\delta }_{PSP}}$ are positively correlated with $\epsilon_{t}$.

\textbf{The attribute similarity threshold $\epsilon_a$.} To assess the impact of the attribute similarity threshold $\epsilon_a$ on model performance, we studied the performance of node classification with various $\epsilon_a$ from 0.2 to 0.9 on all four datasets, as shown in Fig.~\ref{fig:sim}. The Macro-F1 and Micro-F1 increase first and then decrease. This observation indicates that a large threshold may lead to insufficient reconstructed neighbor relationships to obtain informative node embeddings, while a small threshold leads to too many noisy neighbor relationships to weaken information propagation. When the threshold is between 0.5 and 0.7, our method achieves relatively stable and better performance on the four datasets.

\section{Conclusion}\label{sec:conclusion}
We developed a novel and robust contrastive learning approach, named HGCL, for mining and analyzing heterogeneous graphs. HGCL adopts two contrastive views on the guidance of node attributes and graph topologies respectively and integrates and enhances the two views by reciprocally contrastive mechanism. The attribute- and topology-guided views employed different attribute and topology fusion mechanisms, which fully excavated the information and reduced noise interference in the graph. High-quality samples in reciprocal contrast further improved the discriminative power of the model. Extensive experimental results demonstrated the effectiveness of the new HGCL approach over the state-of-the-art methods.

\revise{HGCL aims to design a robust self-supervised model for heterogeneous graphs by capturing effective information in node attributes and graph topology respectively, but it is still not perfect. For example, in this work, in the topology guided view, we only use the attributes of the target type node for message passing, and indirectly utilize the information provided by other types of nodes through meta-path linking, that is, only the nodes at the beginning and end of the meta-path participate in message passing, without considering the node information inside the meta-path. In the future, we plan to explore new message propagation mechanisms to more comprehensively and fully encode the complete information on the meta-path to make our model more powerful. Currently, following the positive and negative sample pair selection mechanisms of most existing methods, we also pre-define positive and negative sample pairs in a two-stage manner. Therefore, we plan to design a more effective positive and negative sample pair selection mechanism so that the model can adaptively and dynamically select better sample pairs to learn node representations in an end-to-end manner. Furthermore, HGCL adopts the classic graph contrastive learning paradigm. In the future, we would also like to explore more self-supervised techniques and apply them effectively to heterogeneous graph analysis.}


\bibliographystyle{ACM-Reference-Format}
\bibliography{main}


\end{document}